\documentclass[11pt]{article}

\usepackage[final]{acl}

\usepackage{times}
\usepackage{latexsym}
\usepackage{graphicx}
\usepackage{booktabs}
\usepackage{multirow}
\usepackage{pifont}
\usepackage{amsmath, amssymb}
\usepackage{enumitem}
\usepackage{microtype}
\usepackage{subcaption}

\usepackage{xcolor}
\usepackage{soul}

\definecolor{lightblue}{rgb}{0.85, 0.95, 1.0} 
\definecolor{mediumblue}{rgb}{0.65, 0.85, 1.0}  
\definecolor{customblue}{rgb}{0.55, 0.75, 1.0}
\definecolor{lightgreen}{rgb}{0.80, 1.00, 0.80}
\definecolor{mediumgreen}{rgb}{0.60, 0.90, 0.60}
\definecolor{darkgreen}{rgb}{0.45, 0.85, 0.45} 
\definecolor{lightpink}{rgb}{1.00, 0.85, 0.90}  
\definecolor{mediumpink}{rgb}{1.00, 0.78, 0.88} 
\definecolor{darkpink}{rgb}{1.00, 0.70, 0.85}  

\newcommand{\hlBone}[1]{{\sethlcolor{lightblue}\hl{#1}}}
\newcommand{\hlBtwo}[1]{{\sethlcolor{mediumblue}\hl{#1}}}
\newcommand{\hlBthree}[1]{{\sethlcolor{customblue}\hl{#1}}}

\newcommand{\hlCone}[1]{{\sethlcolor{lightgreen}\hl{#1}}}
\newcommand{\hlCtwo}[1]{{\sethlcolor{mediumgreen}\hl{#1}}}
\newcommand{\hlCthree}[1]{{\sethlcolor{darkgreen}\hl{#1}}}

\newcommand{\hlGthree}[1]{{\sethlcolor{darkpink}\hl{#1}}}

\usepackage{siunitx}
\title{From Scenes to Elements: Multi-Granularity Evidence Retrieval for Verifiable Multimodal RAG}

\author{
 \textbf{Guanhua Chen\thanks{Equal contribution}},
 \textbf{Chuyue Huang\footnotemark[1]},
 \textbf{Yutong Yao},
 \textbf{Shudong Liu},
 \textbf{Xueqing Song}, \\
 \textbf{Lidia S. Chao},
 \textbf{Derek F. Wong\thanks{Corresponding authors.}}
\\
 NLP\textsuperscript{2}CT Lab, Department of Computer and Information Science, University of Macau
\\
\{nlp2ct.guanhua, nlp2ct.chuyue, nlp2ct.yutong, nlp2ct.shudong, xqsongangie\}@gmail.com\\
\{derekfw, lidiasc\}@um.edu.mo\\
}

\begin{document}
\maketitle

\begin{abstract}
Multimodal Retrieval-Augmented Generation (RAG) systems retrieve evidence at coarse granularities (entire images or scenes), creating a mismatch with fine-grained user queries and making failures unverifiable. We introduce \textbf{GranuVistaVQA}, a multimodal benchmark featuring real-world landmarks with element-level annotations across multiple viewpoints, capturing the partial observation challenge where individual images contain only subsets of entities. We further propose \textbf{GranuRAG}, a multi-granularity framework that treats visual elements as first-class retrieval units through three stages: element-level detection and classification, multi-granularity cross-modal alignment for evidence retrieval, and attribution-constrained generation. By grounding retrieval at the element level rather than relying on implicit attention, our approach enables transparent error diagnosis. Experiments demonstrate that GranuRAG achieves up to 29.2\% improvement over six strong baselines for this task.
\end{abstract}

\section{Introduction}

Multimodal Large Language Models (MLLMs) have substantially advanced visual understanding~\cite{alayrac2022flamingovisuallanguagemodel,li2023blip2bootstrappinglanguageimagepretraining}. However, their reasoning remains opaque and prone to hallucination~\cite{bai2024hallucination}. While Retrieval-Augmented Generation (RAG) mitigates this by conditioning generation on external evidence~\cite{lewis2020retrieval, DBLP:conf/acl/0006YC0W25}, current multimodal extensions operate at coarse granularities, retrieving entire images, scenes, or pages~\cite{fang2024towards,DBLP:conf/iclr/YuTXCRYLWHL025}. This creates a fundamental attribution gap. A user asking about a Baroque pediment may receive a building photograph, yet neither party can verify whether the pediment appears in the image, whether the retrieved knowledge is relevant, or whether the answer faithfully reflects these inputs. Detection failures, retrieval errors, and generation hallucinations collapse into an opaque black box.

In this work, we argue that verifiable multimodal RAG requires treating visual elements as first-class retrieval targets, not merely as regions to be implicitly attended within retrieved scenes. Although recent grounding models can localize objects~\cite{peng2023kosmos,guo2024regiongpt}, they rely on parametric knowledge and do not retrieve external evidence. Conversely, existing multimodal RAG systems retrieve evidence but lack explicit element-level grounding, even recent fine-grained approaches~\cite{liu2025hm} prioritize representational expressiveness over transparent attribution. We bridge this gap with a detect-then-retrieve approach: we first detect candidate visual elements via open-vocabulary detection, then assemble hierarchical evidence spanning element-level descriptions and global context, and finally generate answers attributable to specific visual spans and retrieved passages. This design transforms evaluation from black-box answer assessment into transparent evidence auditing: we can separately diagnose whether the correct elements were detected, whether relevant knowledge was retrieved, and whether generation remained faithful to both.

\begin{table*}[h]
\small
\centering
\begin{tabular}{lcccc}
\toprule
Dataset & Multi-granularity & Fine-grained alignment & Partial entities & RAG \\
\midrule
REAL-MM-RAG \cite{wasserman2025real} & \ding{55} & \checkmark & \ding{55} & \checkmark \\
MMDocIR \cite{dong2025mmdocir} & \checkmark & \ding{55} & \ding{55} & \ding{55} \\
M3DocRAG \cite{cho2024m3docrag} & \ding{55} & \ding{55} & \ding{55} & \checkmark \\
MMLongBench-Doc \cite{ma2024mmlongbench} & \ding{55} & \ding{55} & \ding{55} & \ding{55} \\
SPIQA \cite{pramanick2024spiqa} & \ding{55} & \checkmark & \ding{55} & \ding{55} \\
GranuVistaVQA & \checkmark & \checkmark & \checkmark & \checkmark \\
\bottomrule
\end{tabular}
\caption{Comparison of our dataset with five similar multimodal datasets.}
\label{tab:dataset_comparison}
\end{table*}

However, existing benchmarks inadequately assess multi-granularity alignment as shown in Table \ref{tab:dataset_comparison}. Critically, all overlook the partial observation challenge 
inherent to real-world imagery: photographs capture scenes from varying distances and angles, such that a single image depicts only a subset 
of elements present at a location. This challenge pervades domains from architectural photography to medical imaging and satellite sensing, yet no existing benchmark provides the supervision needed to diagnose element-level detection and retrieval under partial visibility. To address this, we introduce \textbf{GranuVistaVQA}, a benchmark centered on architectural heritage landmarks, a domain where elements have well-defined visual semantics, authoritative knowledge sources exist, and multi-view partial observation naturally arises from real-world photography. The dataset comprises 1,422 images across 71 landmarks, where each view covers only 34\% of annotated elements on average. Crucially, we provide human-verified element visibility labels that enable fine-grained error diagnosis unavailable in prior benchmarks. We further propose GranuRAG, a detect-then-retrieve framework that grounds visible elements via open-vocabulary detection, retrieves hierarchical evidence, and generates attribution-constrained answers. 

Experiments show GranuRAG outperforms strong baselines. Moreover, LLMs fine-tuned on our pipeline's reasoning traces surpass both direct fine-tuning and self-generated chain-of-thought (CoT)~\cite{wei2022chain}, demonstrating that explicit multi-granularity alignment provides a more effective supervision signal.

\section{Related Work}

\paragraph{Benchmarks for Multimodal RAG}
Knowledge-intensive multimodal QA has evolved from answer-only evaluation toward attribution-aware assessment requiring verifiable evidence and localized failure analysis. Early benchmarks~\citep{marino2019okvqavisualquestionanswering,schwenk2022aokvqabenchmarkvisualquestion} established the need for external knowledge without explicit grounding. Follow-up datasets added structured supervision: \textsc{ViQuAE}~\citep{10.1145/3477495.3531753} identifies retrieval as the primary bottleneck; \textsc{InfoSeek}~\citep{chen2023infoseek} and \textsc{Encyclopedic-VQA}~\citep{mensink2023encyclopedicvqavisualquestions} provide section-level evidence, revealing gaps due to unreliable entity-section linking. Document-centric benchmarks~\citep{yu2025visragvisionbasedretrievalaugmentedgeneration,xu2025multigranularityretrievalframeworkvisuallyrich} evaluate at page, region, and document level granularities. Recent work incorporates explicit spatial supervision: \textsc{BBox-DocVQA}~\citep{yu2025bboxdocvqalargescale} grounds answers to semantically coherent regions; \textsc{Toloka VQA}~\citep{ustalov2023tolokavisualquestionanswering} requires bounding boxes for answer-supporting objects; \textsc{VISA}~\citep{ma2024visaretrievalaugmentedgeneration} mandates visual source attribution during generation. However, existing benchmarks lack explicit mappings between visual elements and knowledge entries. GranuVistaVQA addresses this by treating visual elements as core evidence units with element-level knowledge alignment for fine-grained, verifiable attribution.

\paragraph{LLM Methods for Multimodal RAG}
Multimodal RAG has advanced through richer retrieval representations, stronger reranking, and controllable attribution. Unified dense retrievers~\citep{liu2023universalvisionlanguagedenseretrieval,zhou2024marvelunlockingmultimodalcapability,zhou2024vistavisualizedtextembedding} use joint text-image embeddings but lack explicit local visual-textual connections. Finer-grained methods~\citep{lin2024preflmrscalingfinegrainedlateinteraction,yang2025omgmorchestratemultiplegranularities} operate at paragraph/section levels, leaving element-level grounding underexplored. End-to-end pipelines~\citep{DBLP:conf/emnlp/ChenHCVC22,zhang2024mr2agmultimodalretrievalreflectionaugmentedgeneration} improve recall through iterative retrieval but amplify noise. Robustness-focused approaches~\citep{cui2024moremultimodalretrievalaugmented,Yan_2024,tian2025coremmragcrosssourceknowledgereconciliation} address cross-source reconciliation but reason over coarse units. Recent spatial control work targets specific evidence chain components: \textsc{Locate-Then-Generate}~\citep{zhu2023locategeneratebridgingvision} separates localization from generation in scene-text VQA; \textsc{HuLiRAG}~\citep{xi2025tamingretrievalframeworkread} decouples retrieval from attention via segmentation but lacks grounding-knowledge connection; \textsc{GROUNDHOG}~\citep{zhang2024groundhoggroundinglargelanguage} achieves pixel-level alignment without retrieval integration; \textsc{VisRAG 2.0}~\citep{DBLP:journals/corr/abs-2510-09733} improves multi-image reasoning but treats evidence as disconnected region sets; \textsc{Ferret-v2}~\citep{zhang2024ferretv2improvedbaselinereferring} enables fine-grained region-language alignment for in-image comprehension without external knowledge integration. In contrast, GranuRAG treats individual elements as verifiable units, grounding each factual claim to both a detected region and a retrieved snippet, which enables fine-grained alignment and supports systematic error diagnosis across detection, retrieval, and generation stages.

\section{GranuVistaVQA Benchmark}
\label{sec:dataset}

To enable verifiable multimodal RAG with fine-grained attribution, we construct a knowledge-intensive benchmark centered on urban architectural heritage. Unlike prior datasets that treat images as atomic units, our design establishes visual elements as first-class retrieval targets and explicitly models the partial observation challenge: real-world photographs capture landmarks from varying viewpoints, each depicting only a subset of architecturally significant elements.

\subsection{Task Formulation}

Given a query image $I$ depicting a landmark from an arbitrary 
viewpoint, the task is to generate a comprehensive description 
covering all visible architectural elements while avoiding 
hallucination about occluded or absent components.

We associate each landmark with three components: metadata (name, summary, and style) that provides high-level context; an element inventory $E = \{e_1, \ldots, e_k\}$ listing architecturally significant components; and element descriptions $\mathrm{ED}: E \to Paragraphs$ that map each element to expert-written text. For each image $I$, the ground-truth visible set $E^{\mathrm{gt}}(I) \subseteq E$ contains elements visually identifiable in that view. This formulation enables modular evaluation: systems must (i) predict visible elements $\hat{E}(I) \approx E^{\mathrm{gt}}(I)$, (ii) retrieve relevant descriptions from $\mathrm{ED}$, and (iii) generate outputs faithful to both visual and textual evidence. This decomposition allows us to isolate failures at each stage, distinguishing detection errors from retrieval mistakes and generation hallucinations.

\subsection{Data Collection and Annotation}
\paragraph{Domain Selection}
We focus on architectural heritage for three methodological reasons: (1)~elements have well-defined visual semantics amenable to detection, (2)~authoritative knowledge sources enable reliable ground truth, and (3)~tourist photography naturally exhibits multi-view partial observation. We curate 71 landmarks from official cultural heritage databases, spanning religious buildings, temples, fortifications, and cultural institutions across diverse architectural traditions. 

\paragraph{Knowledge Corpus Construction}
For each landmark, we compile a structured JSON document following 
a two-level schema (full specification in Appendix~\ref{app:schema}):
\begin{equation}
    x_{\mathrm{landmark}} = (\mathrm{meta}, E, \mathrm{ED})
\end{equation}
Textual content is sourced from official tourism portals and 
encyclopedic references, then structured through: (i)~element phrase extraction from authoritative descriptions, (ii)~cross-landmark normalization to ensure consistent terminology (e.g., ``bell tower'' $\equiv$ ``campanile''), and (iii)~LLM-assisted description generation with human validation (details in Appendix~\ref{app:pipeline}). Specifically, we focus on heritage sites in Macau, with all description content in Chinese.

\begin{figure}[t]
\centering
\captionsetup[subfigure]{labelformat=parens,labelsep=space,
  justification=centering,singlelinecheck=false,font=footnotesize}

\begin{subfigure}[b]{0.32\linewidth}
  \centering
  \includegraphics[height=2.5cm,keepaspectratio]{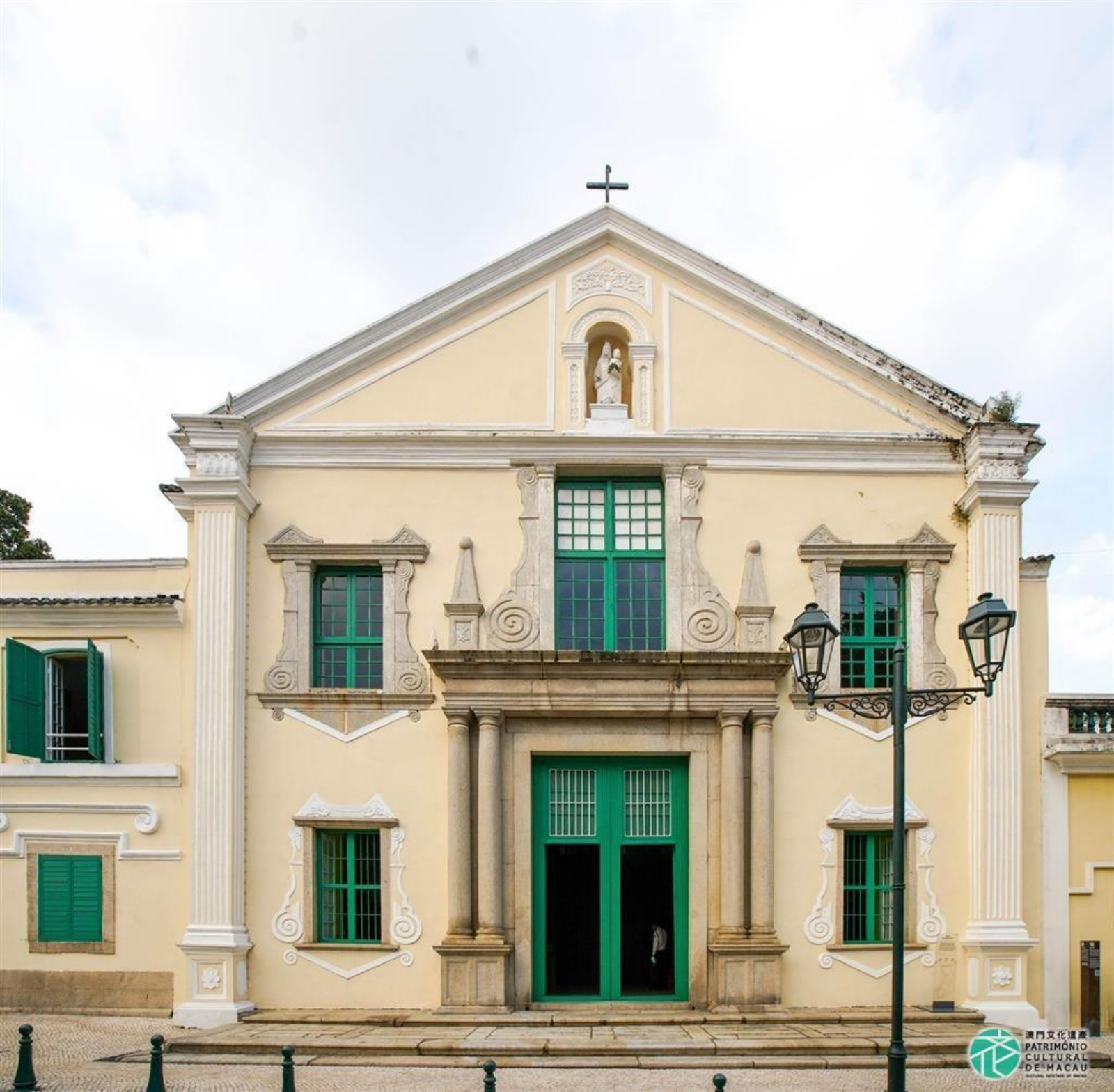}
  \caption{Panoramic}
  \label{fig:view_pano}
\end{subfigure}\hfill
\begin{subfigure}[b]{0.32\linewidth}
  \centering
  \includegraphics[height=2.5cm,keepaspectratio]{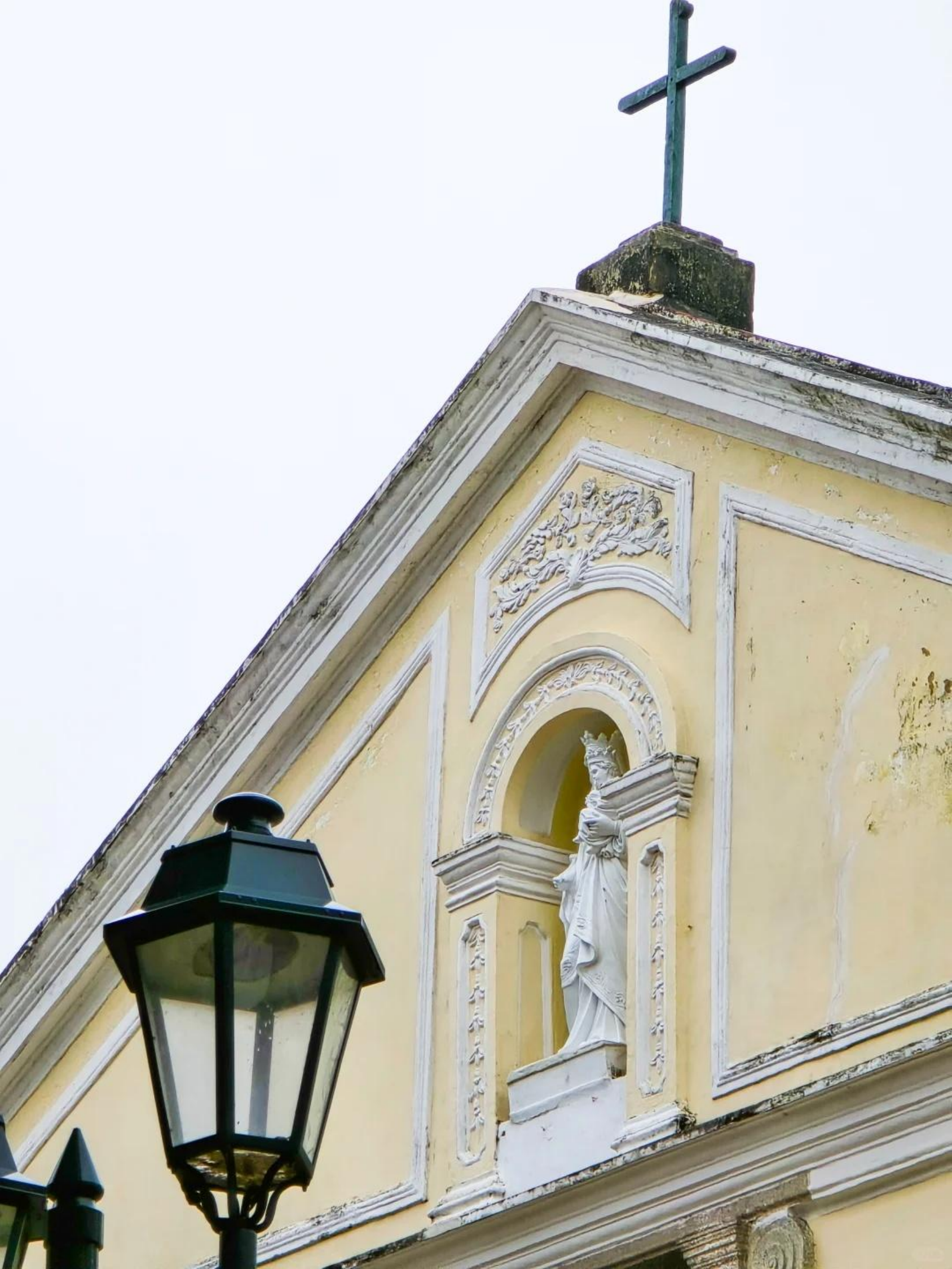}
  \caption{Close-up}
  \label{fig:view_close}
\end{subfigure}\hfill
\begin{subfigure}[b]{0.32\linewidth}
  \centering
  \includegraphics[height=2.5cm,keepaspectratio]{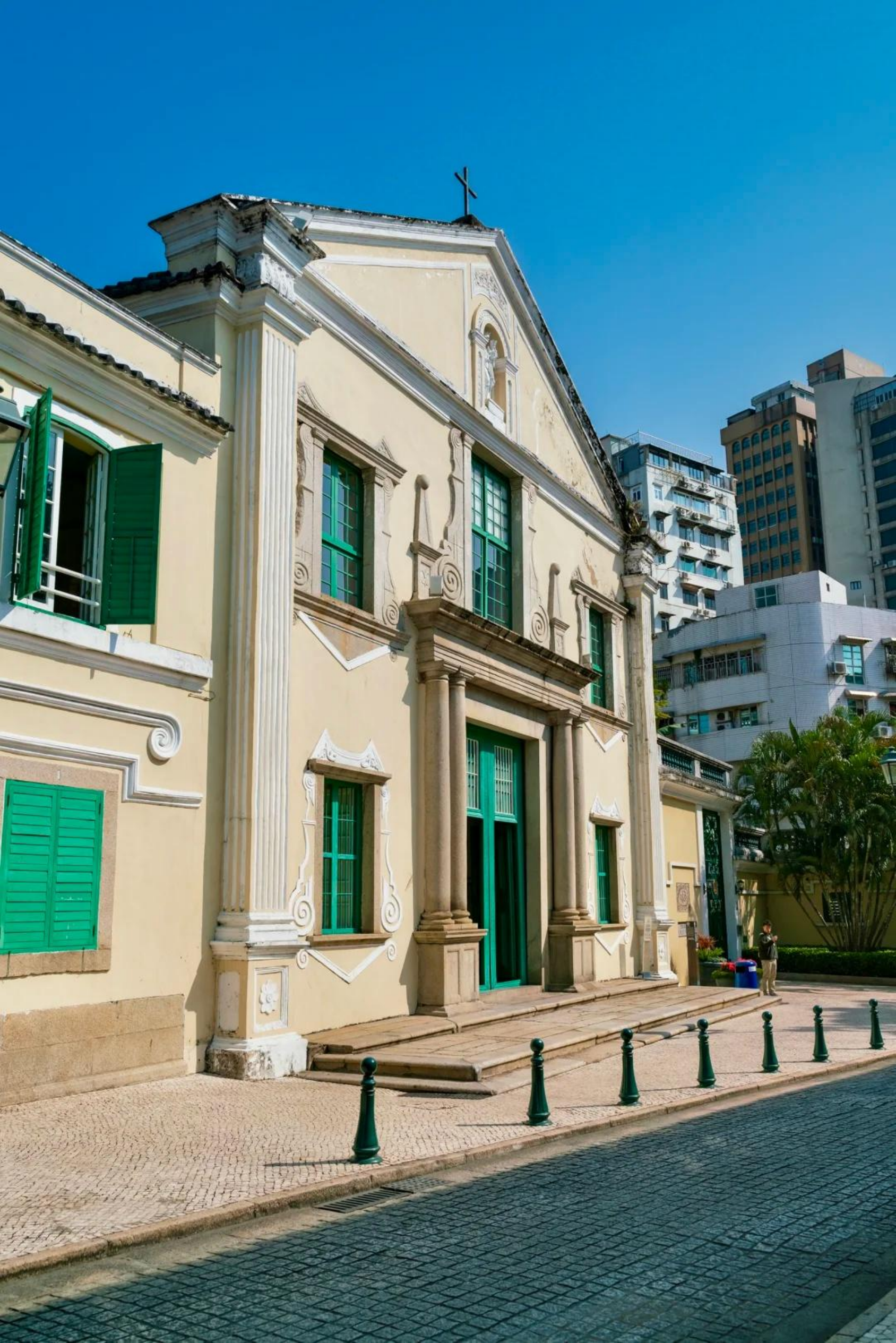}
  \caption{Partial}
  \label{fig:view_part}
\end{subfigure}

\caption{Examples of Multi-Perspective Image.}
\label{fig:view_examples}
\end{figure}

\paragraph{Image Collection}

We collect 1,422 photographs ensuring viewpoint diversity: 
panoramic shots capturing overall structure, close-ups revealing 
fine ornamentation, and oblique partial views (Figure~\ref{fig:view_examples}). After collection, we perform comprehensive data sanitization to remove privacy-sensitive content, including watermarks, visible human faces, and personally identifiable information. We also apply quality filtering to retain only images with resolution $\geq$512px and no visible artifacts. The full screening protocol is described in Appendix~\ref{app:pipeline}.

\paragraph{Visibility Annotation}
For each image $I$, annotators identify $E^{\mathrm{gt}}(I)$ 
following strict visibility criteria:
\begin{itemize}[nosep,leftmargin=*]
    \item \textbf{Visual identifiability}: Elements qualify only 
          if recognizable from pixels alone, without relying on 
          prior landmark knowledge
    \item \textbf{Partial occlusion}: Included only when 
          discriminative visual cues remain (e.g., a half-visible 
          scroll counts if its characteristic shape is apparent)
    \item \textbf{Ambiguity handling}: Uncertain cases are 
          excluded to avoid false positives
\end{itemize}
We employ a human-in-the-loop workflow: an LLM proposes candidate 
elements, which annotators refine by adding missed elements, 
removing hallucinated ones, and resolving synonyms to canonical 
forms (protocol in Appendix~\ref{app:annotation}).

\begin{table}[t]
\centering
\setlength{\tabcolsep}{15pt}
\begin{tabular}{l r}
\toprule
\textbf{Metric} & \textbf{Value} \\
\midrule
\#Landmark ($L$) & 71 \\
\#Img ($N$) & 1422 \\
Avg img/landmark ($N/L$) & 20.03 \\
\#Unique elements ($U_E$) & 221 \\
Avg elements per landmark & 3.59 \\
\bottomrule
\end{tabular}
\caption{The statistic of our proposed GranuVistaVQA.}
\label{tab:stats_corpus}
\end{table}

\begin{figure}[t]
  \centering
  \includegraphics[width=0.95\columnwidth]{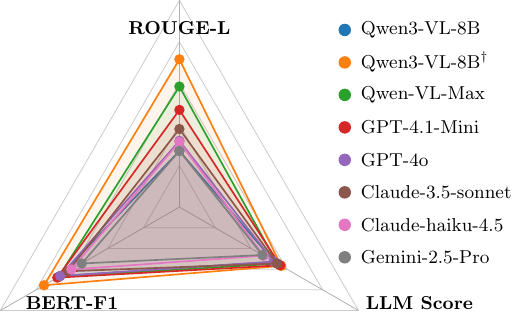}
  \caption{The results of evaluating MLLMs on GranuVistaVQA. \textbf{$^{\dagger}$} means the fine-tuned LLM. }
  \label{fig:radar}
\end{figure}

\begin{figure*}[t]
    \centering
    \includegraphics[width=0.95\linewidth]{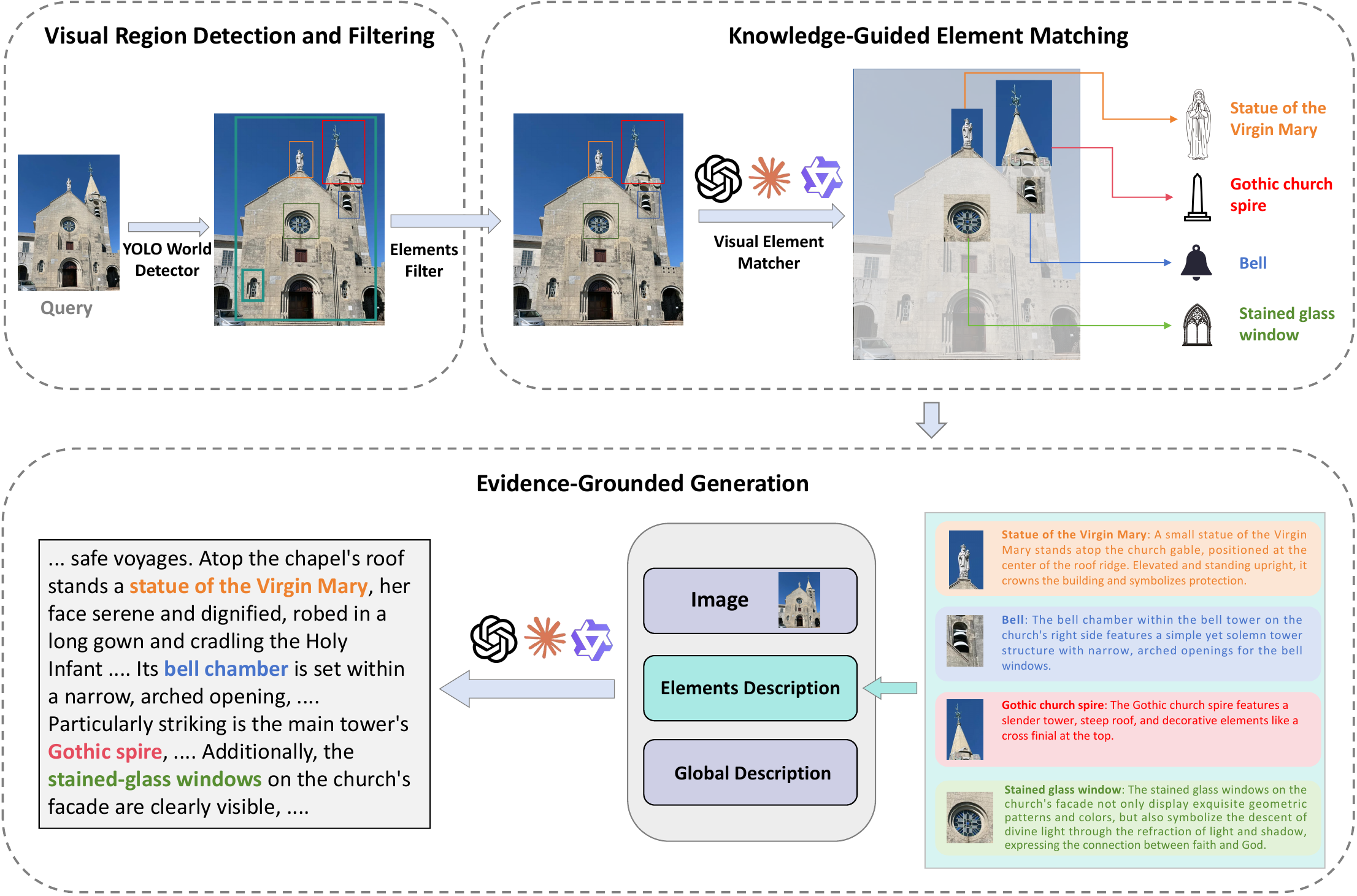}
    \caption{The overview of our proposed GranuRAG framework.}
    \label{fig:mgmmrag_overview}
\end{figure*}

\subsection{Data Statistic and Evaluation} \label{sec:sta&eval}
\paragraph{Statistic} Table~\ref{tab:stats_corpus} summarizes the dataset statistics. The multi-view design creates natural partial observations. close-ups capture fine details but miss broader structures, while panoramic shots show overall layouts but lose granular information (Figure~\ref{fig:view_examples}). On average, individual images cover only ~34\% of their landmark's element inventory. This validates our core premise that answering element-level questions requires aggregating evidence across complementary viewpoints. The statistic of image distribution is explained in Appendix \ref{app:statistics}.

\paragraph{Evaluation Metrics} To evaluate our framework, we report three evaluation metrics: ROUGE-L \cite{lin-2004-rouge} and BERT-F1~\cite{zhang2020bertscoreevaluatingtextgeneration} computed against the gold-standard reference, and an LLM-as-a-judge score \cite{zheng2023judgingllmasajudgemtbenchchatbot}. For the LLM-Score, we use an ensemble of three strong LLMs (GPT-4.1, Gemini-2.5-Pro, claude-haiku-4-5 \cite{2410.21276,google2025gemini,anthropic2025claudehaiku}) to rate each explanation on a $0-100$ scale under a weighted rubric covering Coverage (40\%), Faithfulness (40\%), and Cohesion (20\%); the final LLM Score is the mean across judges to reduce single-model variance \cite{chia2024m}.

\paragraph{Benchmark Evaluation} We evaluated state-of-the-art (SOTA) different powerful MLLMs on our benchmark. Figure~\ref{fig:radar} shows that current models struggle across all metrics as shown in Section \ref{sec:setup}. Even the best-performing model achieves limited success on ROUGE-L and BERT-F1, while all models perform poorly on LLM-as-Judge evaluation, which measures factual accuracy and hallucination control. These results highlight the challenge of multi-granularity information alignment, motivating our GranuRAG framework.

\section{Methodology}
\label{sec:methodology}
Based on the above analysis, we introduce our GranuRAG framework as shown in Figure~\ref{fig:mgmmrag_overview}: given a query image $I$ and candidate element set $E$, we (1)~localize and match visual regions to elements in $E$, yielding visible subset $\hat{E}(I)$; (2)~retrieve hierarchical evidence for $\hat{E}(I)$; and (3)~generate attribution-constrained output grounded in the retrieved evidence and global description.

\subsection{Visual Region Detection and Filtering}
\label{sec:method-benchmark}

The first stage localizes salient architectural regions in image $I$ without requiring prior knowledge of which elements are present. We employ YOLO-World~\cite{cheng2024yolo}, an open-vocabulary object detector, to identify candidate regions based on generic architectural primitives such as columns, carvings, and decorative motifs. This detection strategy offers broad coverage across diverse heritage landmarks without domain-specific fine-tuning.

Raw detections often contain redundant bounding boxes that capture the same visual content at multiple scales. To address this issue, we apply overlap-based filtering: when two boxes overlap by more than 80\%, we retain the smaller one to preserve fine-grained architectural details. The sensitive analysis of this overlap percentage will be shown in Appendix \ref{app.5}. This denoising step produces a refined set of cropped regions:
\begin{equation}
    \mathcal{B}(I) = \{b_1, b_2, \ldots, b_K\}
\end{equation}
where each $b_k$ represents a distinct visual region likely to contain meaningful architectural content. The filtered crops serve as visual queries for the subsequent matching stage.

\subsection{Knowledge-Guided Element Matching}
\label{sec:method-retrieval}

Given the detected regions $\mathcal{B}(I)$ and a candidate element set $E$ with associated appearance descriptions $\{a_e\}_{e \in E}$, this stage determines which architectural elements are actually visible in the image. We formulate this as a multimodal matching problem where each cropped region is compared against element descriptions from the knowledge corpus.

An MLLM receives each annotated bounding box specifically for an image alongside all appearance descriptions, which specify visual attributes such as shape, material, and stylistic features. For each detected region $b_k$, the model identifies the best-matching element by comparing observed visual characteristics against documented descriptions:
\begin{equation}
    e_k = M_\phi\bigl(I, b_k, \{(e, a_e)\}_{e \in E}\bigr) \in E \cup \{\varnothing\}
\end{equation}
where $M_\phi$ is the MLLM parameterized by $\phi$, and the output $\varnothing$ indicates that no candidate description is sufficiently consistent with the region, in which case the region is discarded. This design combines the spatial localization capability of the detector with the fine-grained semantic discrimination ability of the MLLM, enabling reliable identification even among visually similar elements.

The matching process yields a grounded element set $\hat{E}(I) = \{e_k \mid e_k \neq \varnothing\}$ containing only elements with confirmed visual evidence. By explicitly filtering out undetected elements, we establish a principled boundary between what the system observes and what it knows, reducing the risk of hallucinating information about absent components.

\subsection{Evidence-Grounded Generation}
\label{sec:method-generation}

The final stage synthesizes a coherent interpretation by conditioning on both the annotated image and retrieved knowledge for matched elements. For each $e \in \hat{E}(I)$, we retrieve its expert-written description $d_e$ from the knowledge corpus, which provides factual details including historical background, symbolic meaning, and architectural significance. We prepend global metadata $m$ containing landmark name, architectural style, and historical period to contextualize the element-level information:
\begin{equation}
    \mathcal{C}(I) = [m] \oplus \bigl[(e_i, d_{e_i})\bigr]_{e_i \in \hat{E}(I)}
\end{equation}

The generator then produces the final output conditioned on this hierarchical evidence:
\begin{equation}
    y = G_\theta\bigl(I, \mathcal{C}(I) \mid \Omega(\hat{E}(I))\bigr)
\end{equation}
where the generation prompt $\Omega$ instructs the model to describe only elements confirmed in $\hat{E}(I)$ and ground all factual claims to retrieved descriptions. Such a design ensures that each claim in the output traces back to a detected visual region and a retrieved knowledge snippet. Such traceability facilitates systematic error diagnosis: missing information indicates detection failure, incorrect facts suggest retrieval error, and unsupported claims reveal generation hallucination.

\section{Experiment}
\label{sec:exp-setup}

\subsection{Setup} \label{sec:setup}
Experiments are conducted on the GranuVistaVQA dataset as described in section~\ref{sec:dataset}. We use the official APIs of \textbf{Qwen3-VL-8B}, \textbf{Qwen-VL-Max}, \textbf{GPT-4o}, \textbf{GPT-4.1-Mini}, and \textbf{claude-3.5-sonnet} \cite{DBLP:journals/corr/abs-2308-12966,2410.21276,claude35}. For the element detector, we employ the open-vocabulary YOLO-World-XL\footnote{https://replicate.com/franz-biz/yolo-world-xl} $D_\psi$ with a fixed confidence threshold. We use the evaluation metrics following Section \ref{sec:sta&eval}. More implementation details will be shown in Appendix \ref{app.1}.

\begin{table}[t]
\centering
\small
\setlength{\tabcolsep}{5pt} 
\begin{tabular}{lccc}
\toprule
\textbf{Model} & \textbf{ROUGE-L} & \textbf{BERT-F1} & \textbf{LLM} \\
\midrule
Qwen3-VL-8B (A) & 10.88 & 40.04 & 52.90 \\
Qwen3-VL-8B (B) & 12.23 & 44.46 & 61.17\\
Qwen3-VL-8B (C) & \textbf{18.82} & \textbf{46.49} & \textbf{65.00} \\
\midrule
Qwen3-VL-8B$^{\dagger}$ (A) & 28.61 & 45.38 & 56.90 \\
Qwen3-VL-8B$^{\dagger}$ (B)  & 31.96 & 46.05 & 63.20 \\
Qwen3-VL-8B$^{\dagger}$ (C) & \textbf{35.74} & \textbf{46.96} & \textbf{70.24} \\
\midrule
Qwen-VL-Max (A)        & 23.34 & 40.53 & 54.70 \\
Qwen-VL-Max (B)        & 30.29 & 48.39 & 74.10 \\
Qwen-VL-Max (C)        & \textbf{32.01} & \textbf{51.60} & \textbf{83.90} \\
\midrule
GPT-4.1-Mini (A)       & 18.81 & 40.94 & 56.30 \\
GPT-4.1-Mini (B)       & 19.55 & 42.79 & 68.90 \\
GPT-4.1-Mini (C)       & \textbf{22.72} & \textbf{44.04} & \textbf{75.10} \\
\midrule
GPT-4o (A)             & 12.83 & 39.95 & 52.70 \\
GPT-4o (B)             & 18.41 & 42.09 & 63.97 \\
GPT-4o (C)             & \textbf{19.16} & \textbf{43.43} & \textbf{75.40} \\
\midrule
Claude-3.5-Sonnet (A) & 15.11 & 37.26 & 54.50 \\
Claude-3.5-Sonnet (B) & \textbf{16.39} & 39.54 & 66.70 \\
Claude-3.5-Sonnet (C) & 16.36 & \textbf{41.27} & \textbf{80.20} \\
\bottomrule
\end{tabular}
\caption{The main results of our method on four different LLMs under three settings: (A) Baseline, (B) CoT, and (C) GranuRAG. \textbf{$^{\dagger}$} means the fine-tuned LLM. \textbf{Bold} means the best results for each LLM.}
\label{tab:main}
\end{table}

\subsection{Main Results} \label{main-result}
We evaluate GranuRAG across six state-of-the-art MLLMs under three settings: (A) Baseline, where the generator observes the image and the full noisy candidate set $E_{\text{all}}$; (B) CoT, which augments (A) with chain-of-thought prompting~\cite{wei2022chain} to encourage structured reasoning; and (C) GranuRAG, where the generator receives only the grounded element subset $\hat{E}(I)$ from our two-stage pipeline. For both the fine-tuned models (denoted by $^{\dagger}$) and Setting B, we synthesize the thinking process using powerful LLM and validate them through manual inspection (see Appendix~\ref{app.2} for details).

Table~\ref{tab:main} presents the results. Setting C consistently outperforms both baselines across almost all LLMs and metrics, demonstrating that explicit element alignment yields robust improvements regardless of backbone capacity. While CoT prompting (Setting B) moderately improves over the noisy baseline, it does not close the gap to our full pipeline, suggesting that reasoning-time prompting alone cannot fully compensate for irrelevant knowledge. Fine-tuning further amplifies these gains: Qwen3-VL-8B$^{\dagger}$ (C) achieves 35.74 ROUGE-L, 46.96 BERT-F1, and 70.24 LLM score, substantially outperforming its zero-shot counterpart. Notably, even closed-source LLMs benefit markedly from grounded evidence, with Setting C improving LLM scores by 22.7, 18.8, and 25.7 points over Setting A, respectively. These results confirm that our GranuRAG framework provides complementary value beyond model scale or reasoning prompts.

\begin{figure}[t]
    \centering
    \includegraphics[width=0.48\textwidth]{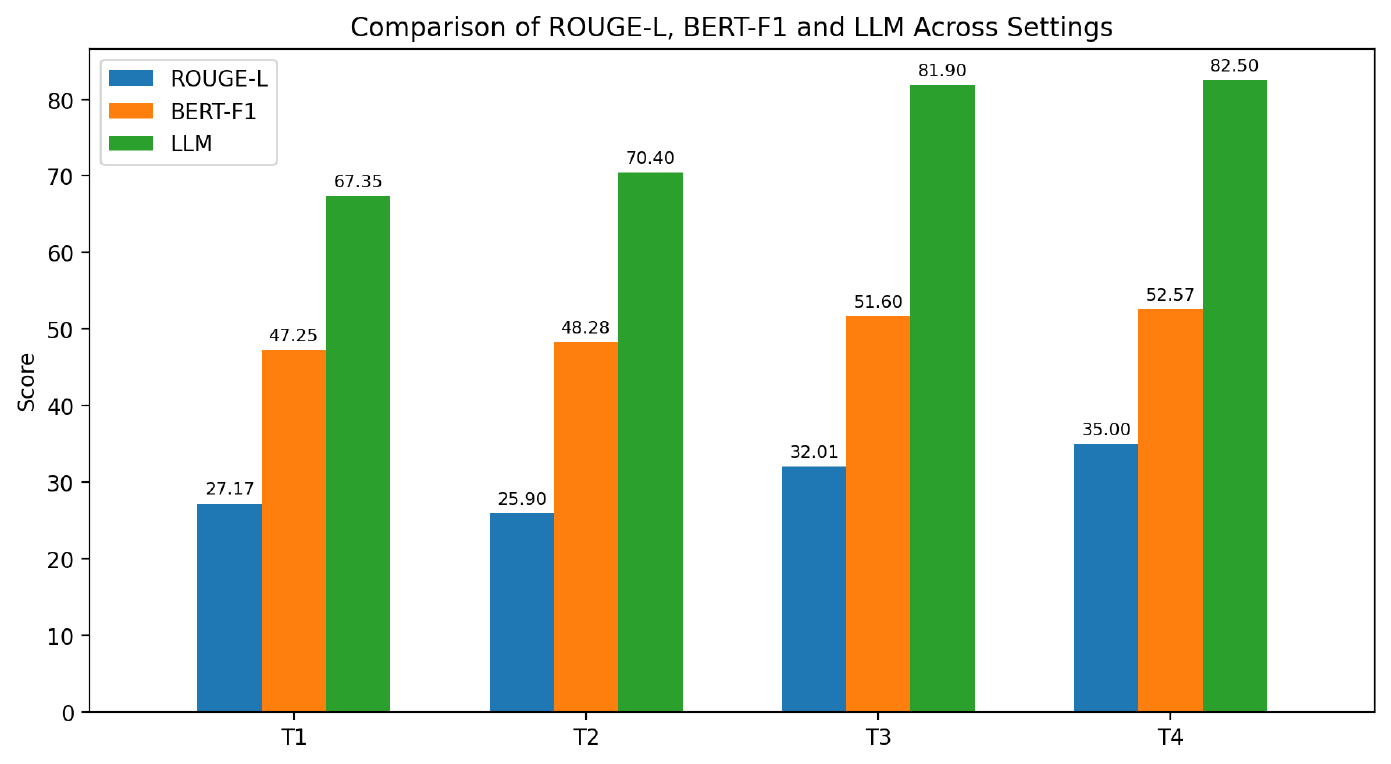}
    \caption{Ablation on visual presentation and element filtering. T3 and T4 use grounded subset $\hat{E}(I)$; T1 and T2 use full candidate set $E_{\text{all}}$.}
    \label{fig:t1_t4_chart}
\end{figure}

\begin{table}[t]
\centering
\small
\setlength{\tabcolsep}{4pt}
\resizebox{\columnwidth}{!}{%
\begin{tabular}{lccc}
\toprule
\textbf{Variant} & \textbf{ROUGE-L} & \textbf{BERT-F1} & \textbf{LLM} \\
\midrule
Text Only (Gold $E_{\text{gold}}$) & \textbf{32.05} & 46.32 & 72.10 \\
Image + All $E_{\text{all}}$ & 29.47 & 45.57 & 68.50 \\
Image + Chosen $\hat{E}(I)$ & 31.01 & \textbf{48.60} & \textbf{74.60} \\
\bottomrule
\end{tabular}%
}
\caption{Ablation on visual modality and knowledge relevance. \textbf{Bold} means the best results.}
\label{tab:ablation_visual_selection}
\end{table}

\subsection{Ablation Studies}
\label{sec:ablation}

We conduct ablation studies to isolate the contributions of visual evidence presentation and knowledge selection. First, we examine four configurations varying visual input and element sets: (T1) raw image with all candidates $E_{\text{all}}$, (T2) box-annotated image with $E_{\text{all}}$, (T3) raw image with grounded subset $\hat{E}(I)$, and (T4) box-annotated image with $\hat{E}(I)$. As shown in Figure~\ref{fig:t1_t4_chart}, T3 and T4 consistently outperform T1 and T2, confirming that filtering noisy candidates is crucial. T4 achieves the highest LLM score, indicating that combining localized visual cues and grounded textual evidence yields optimal performance. 

Table~\ref{tab:ablation_visual_selection} further contrasts three variants: text-only with gold elements $E_{\text{gold}}$, image with all candidates, and image with grounded subset. While text-only provides a strong upper bound, adding the image without filtering (Image + All) paradoxically degrades LLM score, suggesting that irrelevant visual-textual mismatches introduce noise. In contrast, pairing the image with the grounded subset substantially recovers text-only performance and exceeds it on most metrics, demonstrating that selective retrieval enables effective multimodal integration.

We further ablate the detectors to verify that our design choices are well-motivated rather than arbitrary. As shown in Table~\ref{tab:ablation_detector}, replacing YOLO-World with Grounding DINO~\cite{DBLP:conf/eccv/LiuZRLZYJLYSZZ24} still outperforms both the baseline and the approach of directly extracting relevant elements with LLMs, showing that the detect-then-match paradigm is robust across detectors.

\begin{table}[t]
\centering
\small
\setlength{\tabcolsep}{4pt}
\begin{tabular}{lccc}
\toprule
\textbf{Variant} & \textbf{ROUGE-L} & \textbf{BERT-F1} & \textbf{LLM} \\
\midrule
Baseline & 23.34 & 40.53 & 54.70 \\
No detector (LLM-only) & 11.11 & 38.13 & 57.65 \\
Grounding DINO & 27.56 & 41.83 & 72.15 \\
YOLO-World (Ours) & \textbf{32.01} & \textbf{51.60} & \textbf{83.90} \\
\bottomrule
\end{tabular}
\caption{Ablation on the detector component. \textbf{Bold} means the best performance.}
\label{tab:ablation_detector}
\end{table}

\section{Analysis}
\label{sec:main-results}

\subsection{Generalization Analysis}
\label{sec:generalization}
\begin{figure}[t]
    \centering
    \includegraphics[width=0.48\textwidth]{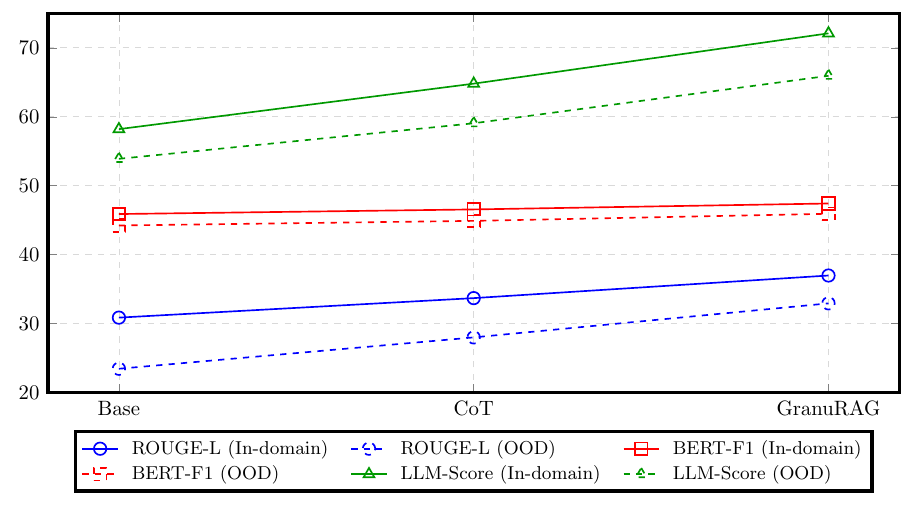}
    \caption{Performance comparison on in-domain and OOD data across three methods.}
    \label{fig:true_false_comparison}
\end{figure}

To assess generalization beyond the fine-tuning distribution, we compare performance on in-domain (ID) and out-of-domain (OOD) test data. ID samples appear in training with different viewpoints, while OOD samples are entirely unseen during training. As shown in Figure~\ref{fig:true_false_comparison}, ID samples consistently achieve higher scores, reflecting expected distributional advantages. Importantly, the absolute improvements from GranuRAG on OOD samples are comparable to, and sometimes exceed, those on ID samples across all three metrics. Moreover, the performance gap between ID and OOD data does not widen as we progress from Base to CoT to GranuRAG. This suggests that our grounding mechanism improves reasoning quality rather than exploiting memorized training instances. The consistent gains on OOD data confirm that GranuRAG generalizes well to novel visual content.

\begin{table}[t]
\centering
\setlength{\tabcolsep}{6pt}
\small
\begin{tabular}{lccc}
\toprule
\textbf{Method} & \textbf{ROUGE-L} & \textbf{BERT-F1} & \textbf{LLM} \\
\midrule
Baseline & 23.79 & 40.83 & 56.40 \\
Embedding Retrieval & 29.47 & 45.57 & 63.45 \\
RAVQA(PreFLMR) & 21.27 & 42.60 & 69.24 \\
VisRAG & 24.06 & 43.35 & 68.06 \\
GranuRAG (Ours) & \textbf{32.27} & \textbf{52.19} & \textbf{79.30} \\
\bottomrule
\end{tabular}%
\caption{The results of comparing our method with different RAG strategies. \textbf{Bold} means the best result.}
\label{tab:main_comparison}
\end{table}

\subsection{Comparison with Retrieval Baselines}

We evaluate the effectiveness of our GranuRAG framework by fixing the generator backbone (\texttt{Qwen-VL-Max}) and comparing three strategies for constructing the evidence that conditions generation. The global baseline provides the generator with descriptions of the full candidate set $E$ without visual grounding, requiring it to implicitly filter irrelevant elements. Embedding Retrieval replaces our second-stage matching with CLIP-based dense retrieval\footnote{https://huggingface.co/openai/clip-vit-large-patch14}~\cite{radford2021learning}: for each detected region, we retrieve the top-1 most similar element description from a vector database. Besides, we also compare GranuRAG with two strong multi-modal RAG frameworks, which is RAVQA~\cite{lin2023fine}\footnote{https://github.com/LinWeizheDragon/Retrieval-Augmented-Visual-Question-Answering} and VisRAG 2.0~\cite{DBLP:journals/corr/abs-2510-09733}\footnote{https://github.com/openbmb/visrag}, using \texttt{Qwen-VL-Max} model.

Table~\ref{tab:main_comparison} shows that GranuRAG substantially outperforms both baselines across all metrics. The global baseline achieves the lowest scores, struggling to identify relevant elements from the full candidate set. Embedding Retrieval improves performance but still lags behind our approach. GranuRAG achieves the best results with particularly large gains in LLM Score (+15.85 over CLIP), demonstrating that LLM-based semantic matching is more effective than embedding similarity for fine-grained element recognition.




\begin{figure}[t]
    \centering
    \begin{subfigure}{0.23\textwidth}
        \centering
        \includegraphics[width=\textwidth]{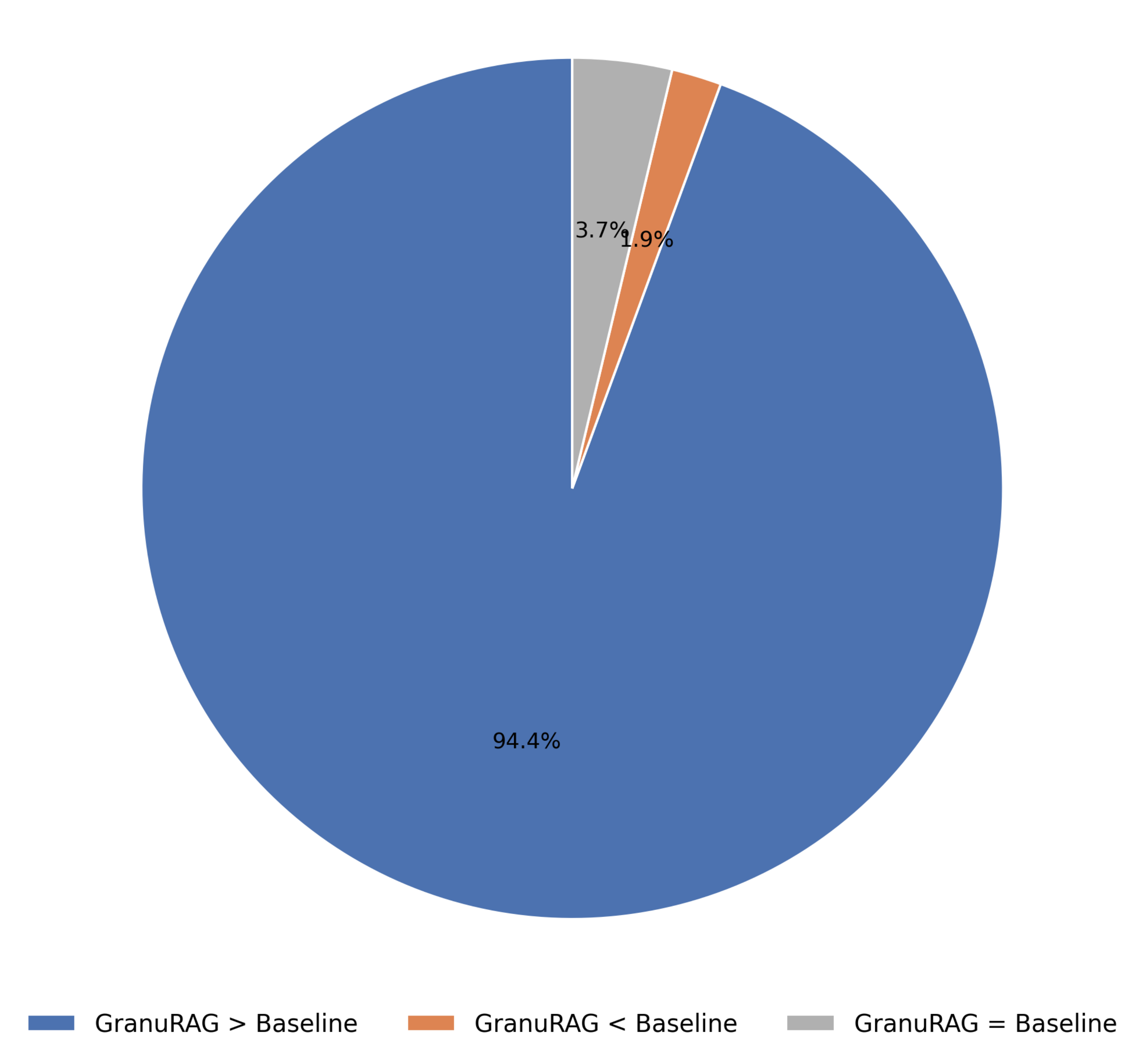}
        \caption{GPT-4o}
        \label{fig:sub-error1}
    \end{subfigure}
    \hfill
    \begin{subfigure}{0.23\textwidth}
        \centering
        \includegraphics[width=\textwidth]{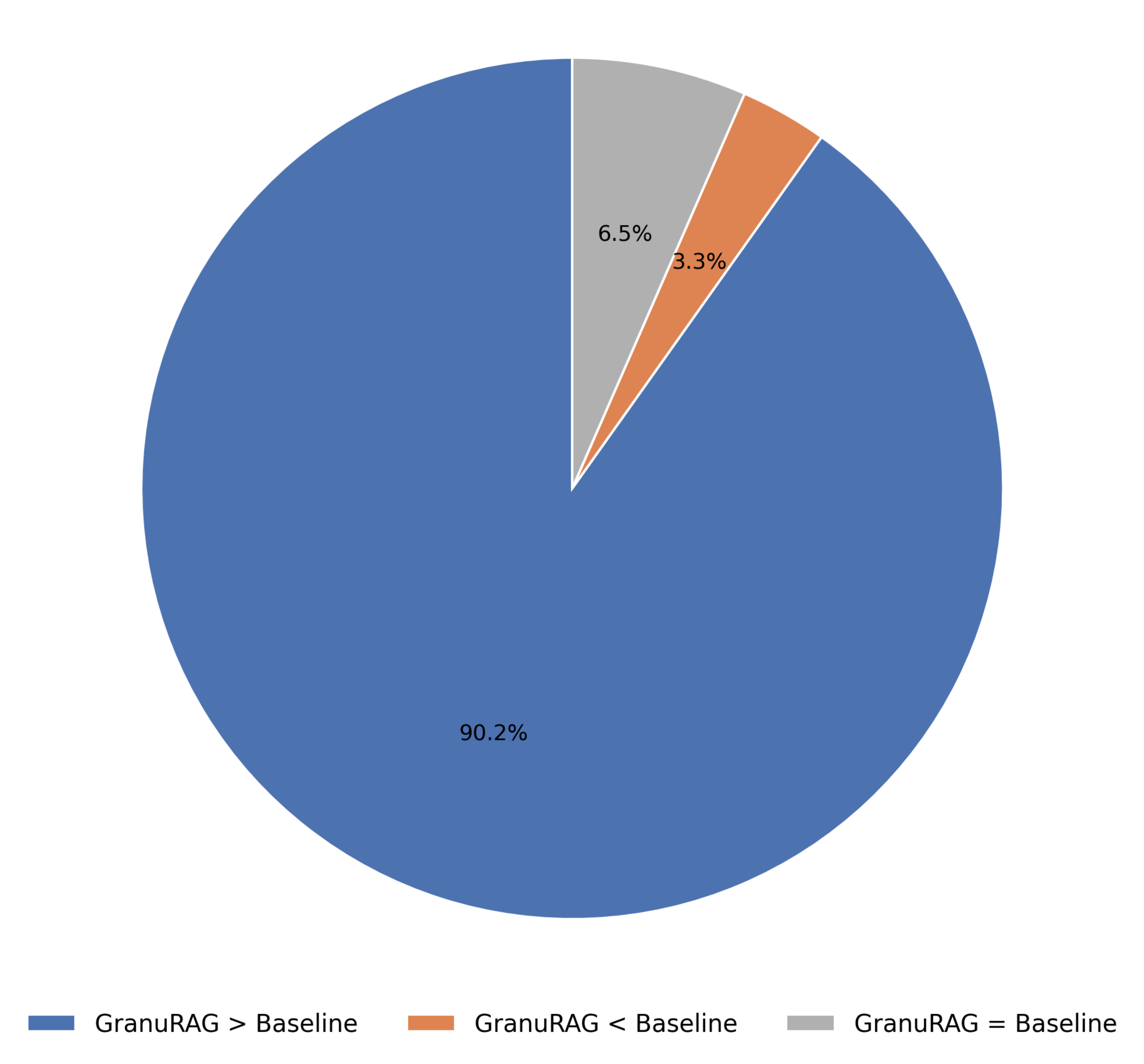}
        \caption{Qwen-VL-Max}
        \label{fig:sub-error2}
    \end{subfigure}
    \caption{Distribution of answer quality when both LLMs successfully extract relevant elements.}
    \label{error1}
\end{figure}

\begin{figure}[t]
    \centering
    \begin{subfigure}{0.23\textwidth}
        \centering
        \includegraphics[width=\textwidth]{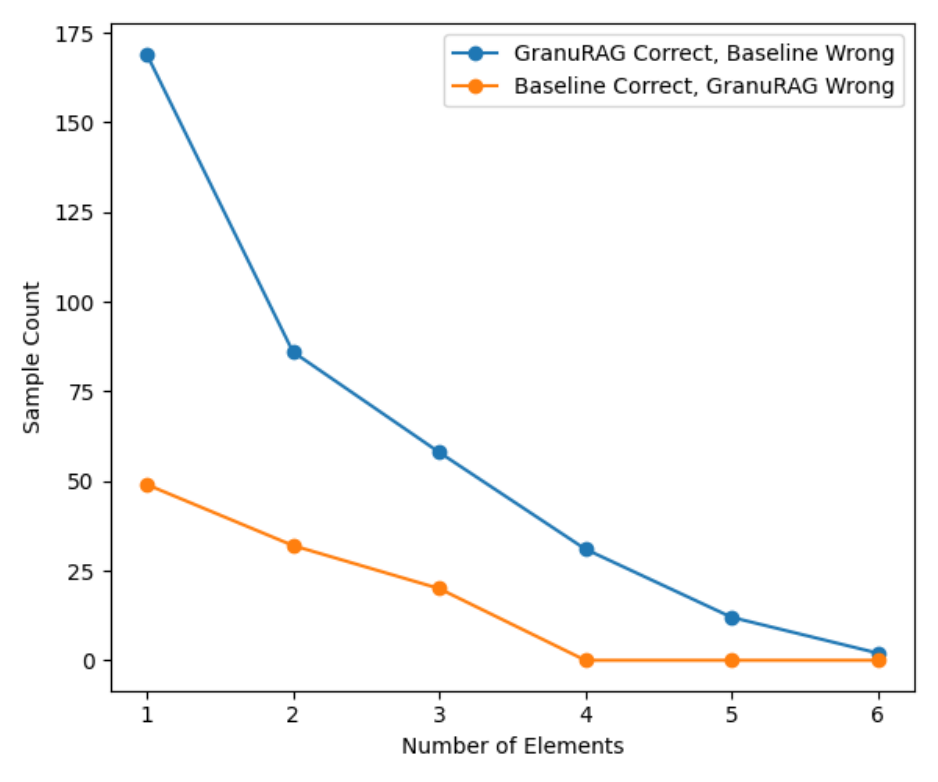}
        \caption{GPT-4o}
        \label{fig:sub-error3}
    \end{subfigure}
    \hfill
    \begin{subfigure}{0.23\textwidth}
        \centering
        \includegraphics[width=\textwidth]{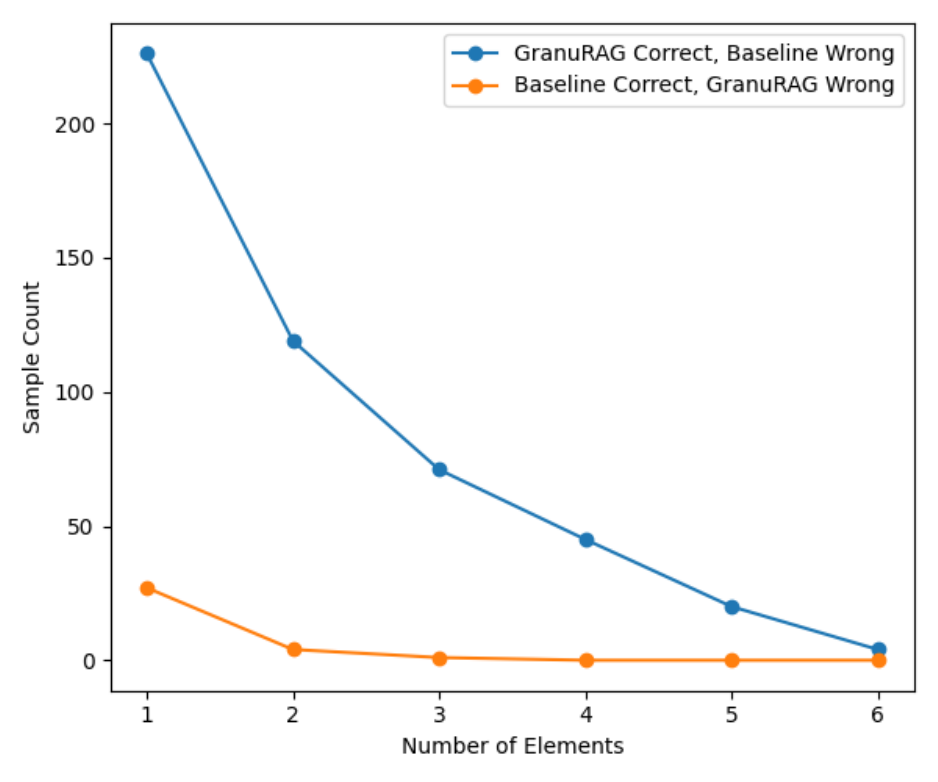}
        \caption{Qwen-VL-Max}
        \label{fig:sub-error4}
    \end{subfigure}
    \caption{Extraction accuracy comparison across images with different numbers of elements.}
    \label{error2}
\end{figure}

\subsection{Error Analysis}

To understand the source of improvements, we conduct error analysis on two LLMs from two aspects. First, we compare answer quality when both methods extract the correct elements. As shown in Figure \ref{error1}, GranuRAG produces better answers in 94.4\% and 90.2\% of cases for the two models respectively, while the baseline wins in less than 5\%. This indicates that our fine-grained element representations provide more accurate information for generation, even when extraction is equally successful.

Second, we analyze extraction accuracy by examining cases where only one method identifies the correct elements. Figure \ref{error2} shows that GranuRAG succeeds where the baseline fails far more often than the reverse, especially for images with fewer elements. This gap narrows as element count increases, reflecting the growing difficulty of fine-grained reasoning in complex images. Overall, these results confirm that GranuRAG improves both extraction accuracy and generation quality.

\begin{table}[t]
\centering
\renewcommand\arraystretch{1}
\tabcolsep=0.45cm    
\begin{tabular}{lcc}
\toprule
\textbf{} & \textbf{win} & \textbf{loss} \\
\midrule
GPT-4o & \textbf{82.22\%} & 17.78\% \\
Qwen-VL-Max & \textbf{91.11\%} & 8.89\% \\
\bottomrule
\end{tabular}%
\caption{The average win and loss rate of human evaluation. \textbf{Bold} means the best results.}
\label{tab:avg_win_loss}
\end{table}

\subsection{Human Evaluation}
\label{sec:human_eval}

To complement automatic metrics, we conduct human evaluation comparing our method against the baseline. We sample 20 questions for two models individually (GPT-4o and Qwen-VL-Max), yielding 40 pairwise comparisons. Three graduate students in computer science independently assess baseline and GranuRAG outputs according to four criteria: \textit{coverage} of relevant scenic elements, \textit{accuracy} of factual details, \textit{fluency} of language, and \textit{acceptability} as tourist guidance. Annotators select the superior output in each pair without knowing which system produced it.

Table~\ref{tab:avg_win_loss} reports the aggregated win rates. GranuRAG achieves 82.22\% preference over the baseline for GPT-4o and 91.11\% for Qwen-VL-Max, confirming that grounded retrieval yields explanations that human evaluators judge more comprehensive and trustworthy. To assess annotation reliability, we compute Fleiss' Kappa across the three raters, obtaining $\kappa = 0.712$ (SE = 0.105, $z = 6.75$, $p < 0.001$), which indicates substantial agreement~\citep{landis1977measurement} and validates the consistency of our evaluation protocol.

\begin{figure}[t]
    \centering
    \begin{subfigure}{0.23\textwidth}
        \centering
        \includegraphics[width=\textwidth]{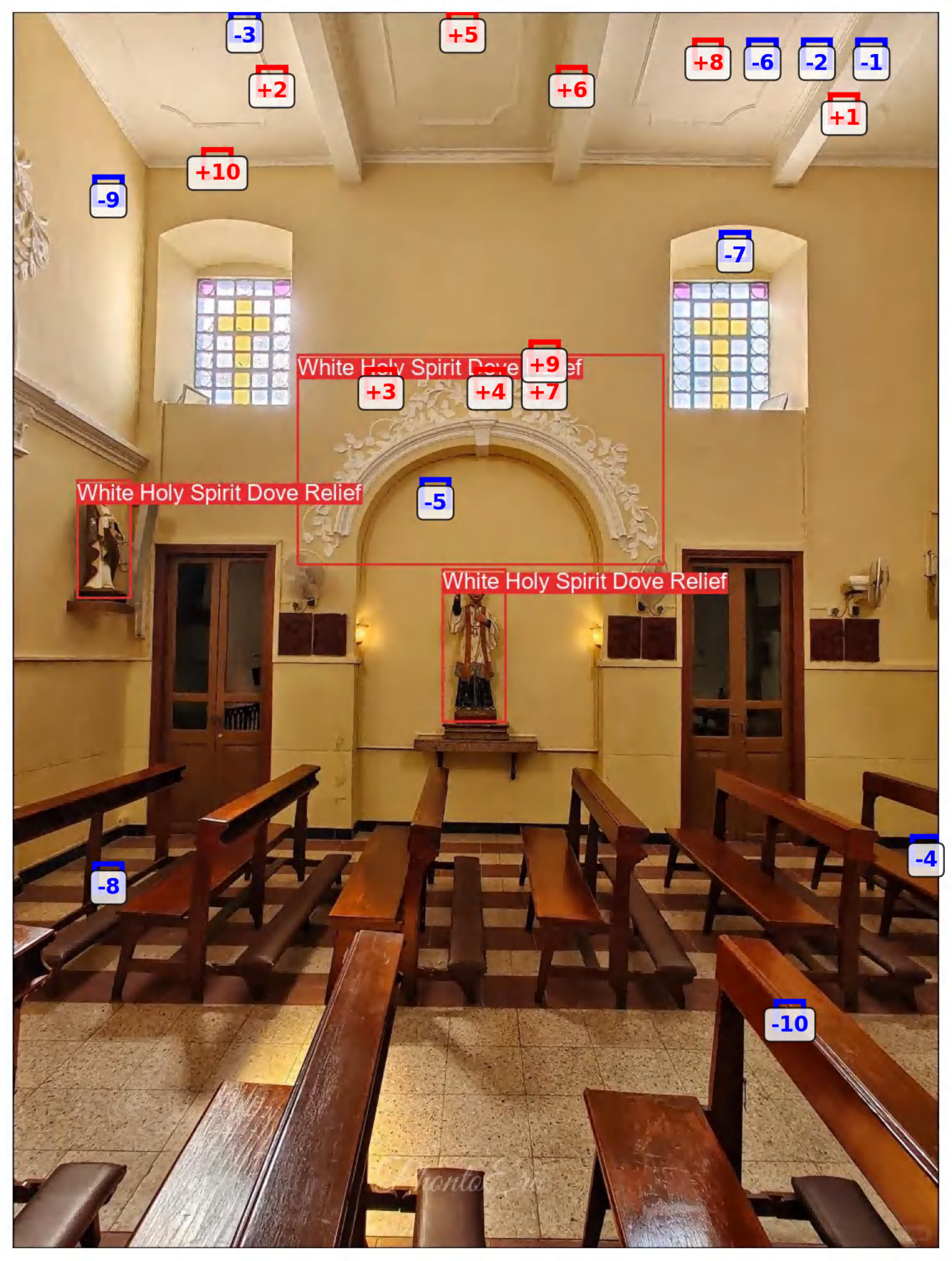}
        \caption{GranuRAG - Base}
        \label{fig:sub-base}
    \end{subfigure}
    \hfill
    \begin{subfigure}{0.23\textwidth}
        \centering
        \includegraphics[width=\textwidth]{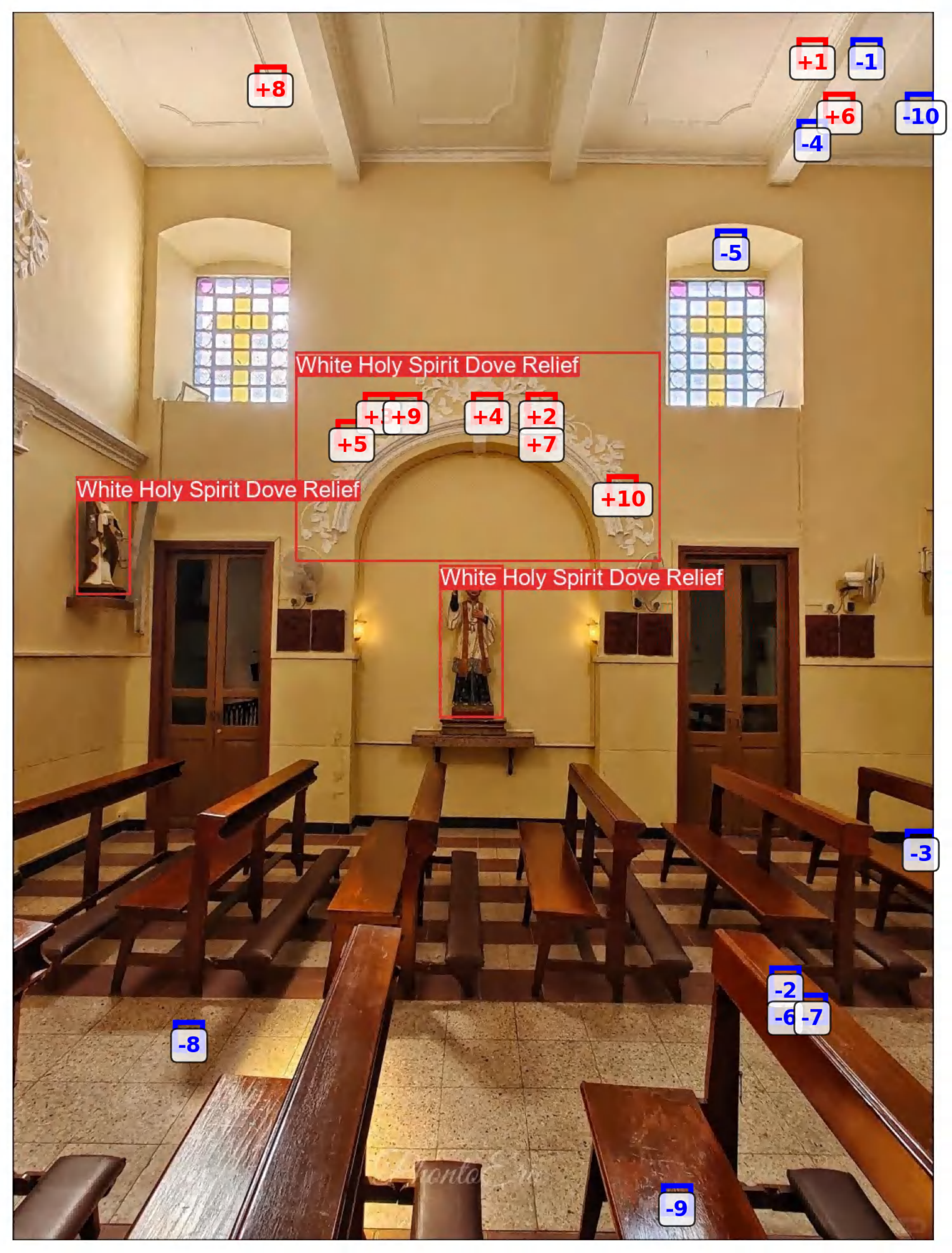}
        \caption{GranuRAG - CoT}
        \label{fig:sub-cot}
    \end{subfigure}
    \caption{Top 10 attention difference regions. \textbf{\color{red}{Red}}: higher attention in our method; \textbf{\color{blue}{Blue}}: higher attention in the base/CoT method.}
    \label{attn-visual}
\end{figure}

\subsection{Visualization}
\label{sec:attention_analysis}

To examine how GranuRAG influences visual focus, we fine-tune three adapters (Base, CoT, and GranuRAG) on Qwen3-VL-8B and compare their attention weights on the same samples. Specifically, we compute the pixel-level difference $\Delta A = A_{\text{GranuRAG}} - A_{\text{baseline}}$ by averaging attention weights across all generated tokens, then visualize the top-10 regions with the largest positive shifts (red $+$ markers) and negative shifts (blue $-$ markers). As shown in Figures~\ref{fig:sub-base} and~\ref{fig:sub-cot}, red markers concentrate on semantically relevant scenic elements, such as the ``White Holy Spirit Dove Relief'' sculptures and decorative arches, while blue markers scatter across generic background regions like ceilings and walls. This pattern confirms that our grounding mechanism systematically reallocates attention toward knowledge-relevant regions and away from distractors. More case studies are provided in Appendix~\ref{app.3}.

\section{Conclusion}
In this paper, we introduced GranuVistaVQA, a benchmark that reflects real-world visual reasoning challenges. Our GranuRAG framework explicitly grounds visual elements before retrieving multi-granularity evidence, enabling transparent attribution and consistently outperforming strong baselines. Experiments demonstrate that organizing evidence hierarchically from atomic visual elements, rather than relying on implicit attention within coarse retrievals, is essential for verifiable multimodal reasoning. 

\section*{Limitations}
While GranuVistaVQA advances multimodal RAG evaluation, several limitations should be noted. First, our benchmark focuses on landmark images where visual elements are typically buildings or monuments. This may not cover all visual reasoning scenarios, such as understanding abstract diagrams or everyday indoor scenes. Second, regarding efficiency, our method takes approximately 3.5 seconds per sample compared to 2 seconds for baseline approaches, nearly twice as long, due to the additional detection and multi-granularity retrieval steps, which may limit applicability in time-sensitive scenarios. Third, our current implementation for the multi-granularity RAG scenario relies solely on traditional supervised fine-tuning (SFT). Future work will explore a broader range of parameter-efficient fine-tuning methods, including diverse LoRA variants \cite{zhang2023adalora, DBLP:journals/corr/abs-2503-23360} as well as recently popular neuron-level fine-tuning techniques~\cite{DBLP:conf/coling/XuZMWC25,DBLP:journals/corr/abs-2603-13201}.

\section*{Acknowledgments}
This work was supported in part by the Science and Technology Development Fund of Macau SAR (Grant Nos. FDCT/0007/2024/AKP, EF2024-00185-FST), the UM and UMDF (Grant Nos. MYRG-GRG2024-00165-FST-UMDF, MYRG-GRG2025-00236-FST), the Tencent AI Lab Rhino-Bird Research Program (Grant No. EF2023-00151-FST), the Stanley Ho Medical Development Foundation (Grant No. SHMDF-AI/2026/001), and the National Natural Science Foundation of China (Grant No. 62266013).

\bibliography{custom}

\appendix
\section{Appendix}
\subsection{Implementation} \label{app.1}
For all evaluated MLLMs, we use a unified decoding setup with temperature set to 0.1 and maximum generation length limited to 600 tokens. This low-temperature setting reduces randomness and ensures stable, reproducible outputs during evaluation. In addition, we fine-tune Qwen3-VL-8B\footnote{https://huggingface.co/Qwen/Qwen3-VL-8B-Instruct} using the LLaMA-Factory\footnote{https://github.com/hiyouga/LLaMA-Factory} framework on a single H800 GPU with setting LoRA~\cite{hu2022lora} rank to 8. The sequence length is set to 4096, with a batch size of 4 and gradient accumulation steps of 4. We train for 3 epochs on the training set with a learning rate of 1e-4. 

\subsection{Sensitivity Analysis of Overlap Threshold} \label{app.5}

We conduct a sensitivity analysis on the overlap threshold used in our redundancy filtering step. Table~\ref{tab:overlap_threshold} reports the results across six threshold values ranging from 70\% to 100\%. As shown, performance follows a clear and intuitive trend: lower thresholds (e.g., 70\%) lead to over-removal of informative elements, which hurts the quality of the generated output, while higher thresholds (e.g., 90\% and above) retain too much redundancy and also degrade performance. The best results are achieved at 80\%, with a ROUGE-L of 32.01, BERT-F1 of 51.60, and LLM Score of 83.90. Importantly, performance remains consistently strong across the 75\%--85\% range, indicating that our method is not overly sensitive to this hyperparameter. We select 80\% as our default threshold since it sits at the natural midpoint of this effective range and yields the best overall performance.

\begin{table}[t]
\centering
\small
\begin{tabular}{cccc}
\toprule
\textbf{Threshold} & \textbf{ROUGE-L} & \textbf{BERT-F1} & \textbf{LLM Score} \\
\midrule
70\%  & 23.69 & 39.35 & 68.00 \\
75\%  & 29.83 & 47.72 & 75.35 \\
80\%  & \textbf{32.01} & \textbf{51.60} & \textbf{83.90} \\
85\%  & 31.67 & 50.02 & 79.30 \\
90\%  & 29.84 & 48.69 & 76.58 \\
100\% & 26.45 & 43.67 & 65.20 \\
\bottomrule
\end{tabular}
\caption{Sensitivity analysis of the overlap threshold. \textbf{Bold} means the best performance.}
\label{tab:overlap_threshold}
\end{table}

\subsection{Synthesize Fine-tuned Data} \label{app.2}
For the fine-tuning task, we construct our test set by randomly selecting 20\% of partial landmarks images (containing incomplete visual elements), with the remainder forming the training set. Critically, we ensure that approximately half of the scenic spots are entirely absent from the training set to evaluate the model's robustness on unseen locations. The related analysis is shown in section \ref{sec:generalization}. We then employ Qwen3-VL-235B-A22B-Thinking~\cite{qwen3technicalreport} to synthesize reasoning data for both sets by populating the prompt with three components: the landmark image, annotated visual elements with descriptions, and the target commentary text. We prompt the LLM to generate reasoning processes from input to target output using two approaches: its native chain-of-thought style and our proposed pipeline framework.

To ensure data quality, each synthesized reasoning process underwent automated scoring by LLM, with samples failing to fully meet our criteria being regenerated. When a sample failed validation three consecutive times, human annotators intervened to manually revise the reasoning process. Through this rigorous pipeline combining automated synthesis with quality control, we construct a high-quality dataset for fine-tuning LLMs to learn both standard CoT reasoning and our pipeline-based thinking process.

\subsection{Attribution Evaluation}
\label{sec:attribution}

Beyond factual correctness, we also evaluate whether LLMs can trace their claims back to specific visual evidence. A description may sound plausible yet be entirely ungrounded in what the model actually perceives. We measure this with three metrics: Attribution Precision (AP), the fraction of cited elements that exist in the predicted visible set $\hat{E}(I)$; Attribution Recall (AR), the fraction of elements in $\hat{E}(I)$ that are actually cited in the output; and Unsupported Claim Rate (UCR), the fraction of output sentences not grounded by any element.

\begin{table}[t]
\centering
\setlength{\tabcolsep}{10pt}
\begin{tabular}{@{}l ccc@{}}
\toprule
\textbf{Model} & \textbf{AP\,$\uparrow$} & \textbf{AR\,$\uparrow$} & \textbf{UCR\,$\downarrow$} \\
\midrule
\multicolumn{4}{@{}l}{\textit{Qwen3-VL-8B}} \\
\quad Setting (A) & 0.6997 & 0.2244 & 0.9417 \\
\quad Setting (B) & 0.8153 & 0.2423 & 0.7814 \\
\quad Setting (C) & \underline{0.9242} & \underline{0.4467} & \underline{0.6753} \\
\midrule
\multicolumn{4}{@{}l}{\textit{Qwen3-VL-8B$^\dagger$}} \\
\quad Setting (A) & 0.6500 & 0.3921 & 0.8392 \\
\quad Setting (B) & 0.8420 & 0.4309 & 0.7513 \\
\quad Setting (C) & \underline{0.9561} & \underline{0.5847} & \underline{0.6235} \\
\midrule
\multicolumn{4}{@{}l}{\textit{Qwen-VL-Max}} \\
\quad Setting (A) & 0.7675 & 0.3573 & 0.7779 \\
\quad Setting (B) & 0.9474 & 0.4754 & 0.6279 \\
\quad Setting (C) & \underline{0.9730} & \underline{0.6194} & \underline{0.5293} \\
\midrule
\multicolumn{4}{@{}l}{\textit{GPT-4.1-Mini}} \\
\quad Setting (A) & 0.6791 & 0.3339 & 0.8104 \\
\quad Setting (B) & 0.9614 & 0.5844 & 0.7888 \\
\quad Setting (C) & \underline{0.9865} & \underline{0.6144} & \underline{0.6482} \\
\midrule
\multicolumn{4}{@{}l}{\textit{GPT-4o}} \\
\quad Setting (A) & 0.6990 & 0.3562 & 0.8490 \\
\quad Setting (B) & 0.9505 & 0.6097 & 0.7323 \\
\quad Setting (C) & \textbf{\underline{0.9865}} & \textbf{\underline{0.6620}} & \textbf{\underline{0.6184}} \\
\midrule
\multicolumn{4}{@{}l}{\textit{Claude-3.5-Sonnet}} \\
\quad Setting (A) & 0.7616 & 0.2219 & 0.9414 \\
\quad Setting (B) & 0.9479 & 0.2494 & 0.8125 \\
\quad Setting (C) & \underline{0.9583} & \underline{0.4297} & \underline{0.6931} \\
\bottomrule
\end{tabular}
\caption{Attribution evaluation results. AP = Attribution Precision, AR = Attribution Recall, UCR = Unsupported Claim Rate. $^\dagger$ denotes the fine-tuned variant. \textbf{Bold} marks the best score per metric. \underline{Underline} means the best results for each LLM.}
\label{tab:attribution}
\end{table}

As shown in Table~\ref{tab:attribution}, all models improve consistently from Setting~(A) to~(C), confirming that structured evidence grounding helps anchor generated text to perceived visual content. Fine-tuning also brings clear gains: Qwen3-VL-8B$^\dagger$ improves recall from 0.45 to 0.58 under Setting~(C) compared to its base version. Among proprietary models, GPT-4o achieves the best overall profile, while Claude-3.5-Sonnet shows high precision but low recall, suggesting it cites carefully but sparingly. Notably, even the best configuration leaves over half of output sentences unsupported, indicating that faithful visual attribution remains an open challenge.

\subsection{Case Study} \label{app.3}

Table~\ref{case study} presents a color-coded case study that directly compares element-level grounding and description quality across methods. Each color family corresponds to one architectural element, with darker shades indicating more accurate descriptions. Specifically, \textcolor{blue}{blue} highlights track the Christian emblem, \textcolor{green}{green} highlights track windows, and \textcolor{pink!60!red}{pink} denotes the pediment. This visualization reveals clear differences: the baseline misses the pediment entirely, CoT hallucinates non-existent twin towers, while GranuRAG stays faithful to the visible evidence and aligns closely with the ground truth.

From an element selection perspective, the baseline fails to identify the prominent pediment, resulting in incomplete structural coverage. CoT, though more detailed, introduces twin towers that do not exist in the image, a typical hallucination risk in CoT reasoning. GranuRAG correctly identifies all key elements including the pediment and Christian emblem while avoiding both omissions and hallucinations. In terms of generation quality, the shade progression reflects meaningful differences: for the Christian emblem (\textcolor{blue}{blue}), GranuRAG provides fine-grained symbolic details (e.g., three nails, IHS inscription) that match the ground truth, while baseline and CoT remain superficial. For windows (\textcolor{green}{green}), GranuRAG delivers complete structural descriptions versus partial coverage by other methods. Most notably, the pediment (\textcolor{pink!60!red}{pink}) appears only in GranuRAG's output, demonstrating that our method improves both element recall and description accuracy.

\renewcommand{\arraystretch}{1.25}





\section{Dataset Construction Details}
\label{app:dataset}

\subsection{Data Schema}
\label{app:schema}

We organize knowledge at two granularities to support hierarchical retrieval:

\noindent\textbf{Landmarks -level representation:}
\begin{equation}
    x_{\mathrm{attr}} = (\mathrm{meta}, E, \mathrm{ED}, I^{\mathrm{rep}}),
\end{equation}
where $\mathrm{meta} = \{\text{Landmarks, Summary, Style}\}$ contains high-level context; $E = \{e_1, \ldots, e_k\}$ is the landmarks-specific element inventory; $\mathrm{ED}: E \to \text{Paragraphs}$ maps each element to its detailed description; and $I^{\mathrm{rep}}$ is a representative image for visualization.

\noindent\textbf{Image-level representation:}
\begin{equation}
    z = (I, a(I), E^{\mathrm{gt}}(I)),
\end{equation}
where $I$ is an image, $a(I)$ denotes its source landmarks, and $E^{\mathrm{gt}}(I) \subseteq E$ specifies ground-truth visible elements.

\noindent The structured JSON schema follows:
\begin{verbatim}
{
  "Landmarks": "[NAME]",
  "Summary": "[description]",
  "Style": "[style_notes]",
  "Elements": ["[elem_1]", ..., 
    "[elem_k]"],
  "ElementDescriptions": {
    "[elem_i]": "[paragraph]", 
    ...
  },
  "Image": "[image_path]"
}
\end{verbatim}

\subsection{Construction Pipeline}
\label{app:pipeline}

\paragraph{Stage 1: Data Collection}
We collect textual materials from encyclopedic sources 
including Wikipedia and Baidu Baike, as well as from 
official heritage and tourism portals such as the Macao 
World Heritage website\footnote{\url{https://www.wh.mo/cn/}} 
and the Macao Government Tourism Office 
website.\footnote{\url{https://www.macaotourism.gov.mo/zh-hant/sightseeing}} 
Images are sourced from Google Images, Rednote, and 
Trip.com using landmark names as search queries to ensure 
viewpoint diversity. All data are collected for 
non-commercial research purposes only. In addition, we perform data sanitization to remove privacy-sensitive and potentially harmful content, employing a hybrid approach that combines LLM-based filtering with manual spot-checking. 
All images are collected exclusively for non-commercial academic research under the principle of fair use. 

We perform comprehensive data sanitization through a rigorous manual screening process. Trained annotators inspect each image individually following a strict exclusion protocol: any image containing the following elements is immediately discarded and not 
included in the dataset:
\begin{itemize}[nosep,leftmargin=*]
    \item Visible watermarks, platform logos, or copyright overlays
    \item Timestamps, date stamps, or camera metadata overlays
    \item Geolocation tags, map overlays, or address information
    \item License plates, vehicle identifiers, or transportation tags
    \item Identity documents, tickets, receipts, or personal contact info
    \item Recognizable human faces and personal information
\end{itemize}
This policy prioritizes privacy protection over 
data retention. After manual screening, we apply quality filtering 
to retain only images with resolution $\geq$512px and no visible 
compression artifacts. The final dataset contains 1,422 images that have passed both privacy and quality checks, ensuring suitability for academic research purposes.

\paragraph{Stage 2: Knowledge Structuring}
Raw textual materials are consolidated into the normalized JSON schema through a four-step process. We first extract element phrases from authoritative descriptions, then apply normalization rules (Appendix~\ref{app:normalization}). Next, we generate element descriptions via LLM (Appendix~\ref{app:ed_generation}) and finally validate all outputs (Appendix~\ref{app:validation}).

\paragraph{Stage 3: Image Annotation}
For each image $I$, we provide Qwen3-VL-Max~\cite{qwen3technicalreport} with the element inventory $E$ and descriptions $\mathrm{ED}$ as context. The model proposes candidate visible elements $\hat{E}(I) \subseteq E$, which human annotators then refine into final ground-truth labels $E^{\mathrm{gt}}(I)$ following strict visibility guidelines detailed in Appendix~\ref{app:annotation}.

\paragraph{Stage 4: Indexing}
We embed all images using CLIP and L2-normalize the resulting vectors to 512 dimensions. A FAISS flat L2 index is built with aligned metadata that maps vector IDs to landmark names, image paths, and ground-truth element sets, enabling source attribution during retrieval.

\begin{table}[t]
\centering
\begin{tabular}{lccc}
\toprule
\textbf{Model} & \textbf{P} & \textbf{R} & \textbf{F1} \\
\midrule
Qwen3-VL-8B          & 0.723 & 0.493 & 0.470 \\
Qwen-VL-Max          & \textbf{0.850} & \textbf{0.822} & \textbf{0.807} \\
GPT-4.1-Mini          & 0.801 & 0.715 & 0.731 \\
GPT-4o                & 0.699 & 0.662 & 0.649 \\
Claude-3.5-Sonnet     & 0.754 & 0.741 & 0.758 \\
\bottomrule
\end{tabular}
\caption{Element detection results with normalization applied. \textbf{Bold} means the best result.}
\label{tab:element_detection}
\end{table}

\subsection{Normalization Procedures}
\label{app:normalization}

We apply three normalization strategies to ensure cross-landmark consistency. Language unification consolidates multilingual variants into canonical keys and enforces consistent orthography. Synonym collapsing merges semantically equivalent terms under lightweight taxonomy tags, such as grouping ``bell tower'' with ``campanile,'' ``granite stone'' with ``granite,'' and ``carved motif'' with ``ornamental carving.'' Cross-landmark alignment ensures that shared architectural features like ``Baroque fa\c{c}ade'' receive identical identifiers across different landmarks. These steps are critical for two reasons: they prevent the element matching stage (Section~\ref{sec:method-retrieval}) from failing due to superficial lexical mismatches across knowledge sources, and they enable fair evaluation by ensuring the same element is not counted as two different entities, which would artificially deflate Precision and Recall. As shown in Table~\ref{tab:element_detection}, even with normalization applied, element detection remains challenging, confirming that the task difficulty is genuine rather than an artifact of inconsistent naming.

\begin{table*}[ht]
\centering
\small
\begin{tabular}{p{0.33\textwidth} p{0.33\textwidth} p{0.33\textwidth}}
\toprule

\multicolumn{3}{l}{\textbf{Input}} \\

\multicolumn{3}{l}{
\parbox{\textwidth}{
Please carefully observe the visible details in the image corresponding to these elements, focusing your description on what is actually visible in the image. Below is the complete element description (ED) for the image.\\

\textbf{ED:}\\
\textbf{Twin Towers:} The top floor features symmetrical bell towers on both sides...\\
\textbf{Windows:} Smaller side windows with arched lintels and surrounding relief decorations...\\
\textbf{Pediment:} The central pediment is ...\\
\textbf{Christian Emblem:} Below the emblem are three nails commemorating...
}
} \\

\midrule

\begin{minipage}[t]{\linewidth}
\centering
\textbf{Original Image}\par\vspace{2pt}
\includegraphics[width=0.9\linewidth]{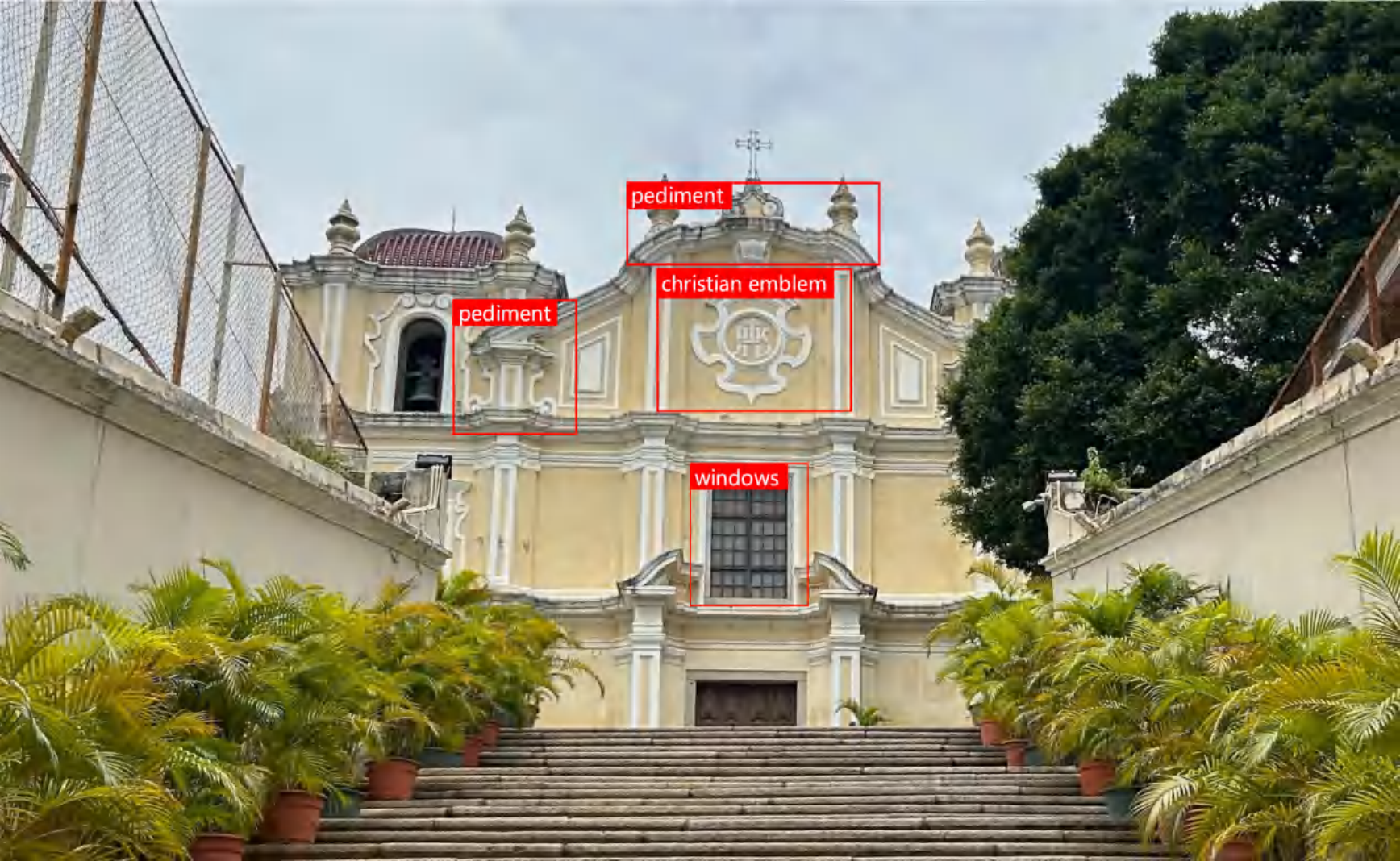}
\end{minipage}
&
\multicolumn{2}{>{\raggedright\arraybackslash}p{0.66\textwidth}}{%
\begin{minipage}[t]{\linewidth}
\centering\textbf{GT Text}\par\vspace{2pt}\raggedright
The facade of this church spans 24.6 meters in width. Its pediment displays quintessential Baroque style, adorned centrally with an exquisite Jesuit coat of arms. This christian emblem, rendered in symmetrical geometric patterns and relief carvings, serves as a common religious symbol in Jesuit churches. Directly below the coat of arms, three nails symbolize the historical events of Jesus' crucifixion. Above the coat of arms, the letters ``IHS'' are carved, derived from the initials of Jesus' Greek name ``$\mathrm{IHSOUS}$,'' commonly used as an abbreviation for Jesus Christ, signifying ``Jesus is the Savior of mankind.'' The pilasters flanking the coat of arms are adorned with broken pediments. The central section of the church's second level features a large rectangular window with louvered panels, flanked by smaller arched windows. The lintels are decorated with scrollwork and complemented by carved details. This layout aligns with the Baroque architectural pursuit of symmetry and opulence.
\end{minipage}
} \\

\midrule

\multicolumn{1}{c}{\textbf{Baseline}} &
\multicolumn{1}{c}{\textbf{CoT}} &
\multicolumn{1}{c}{\textbf{GranuRAG}} \\

\midrule

\multicolumn{1}{l}{\textbf{Annotated Image}} &
\multicolumn{1}{l}{\textbf{Annotated Image}} &
\multicolumn{1}{l}{\textbf{Annotated Image}} \\

\includegraphics[width=\linewidth]{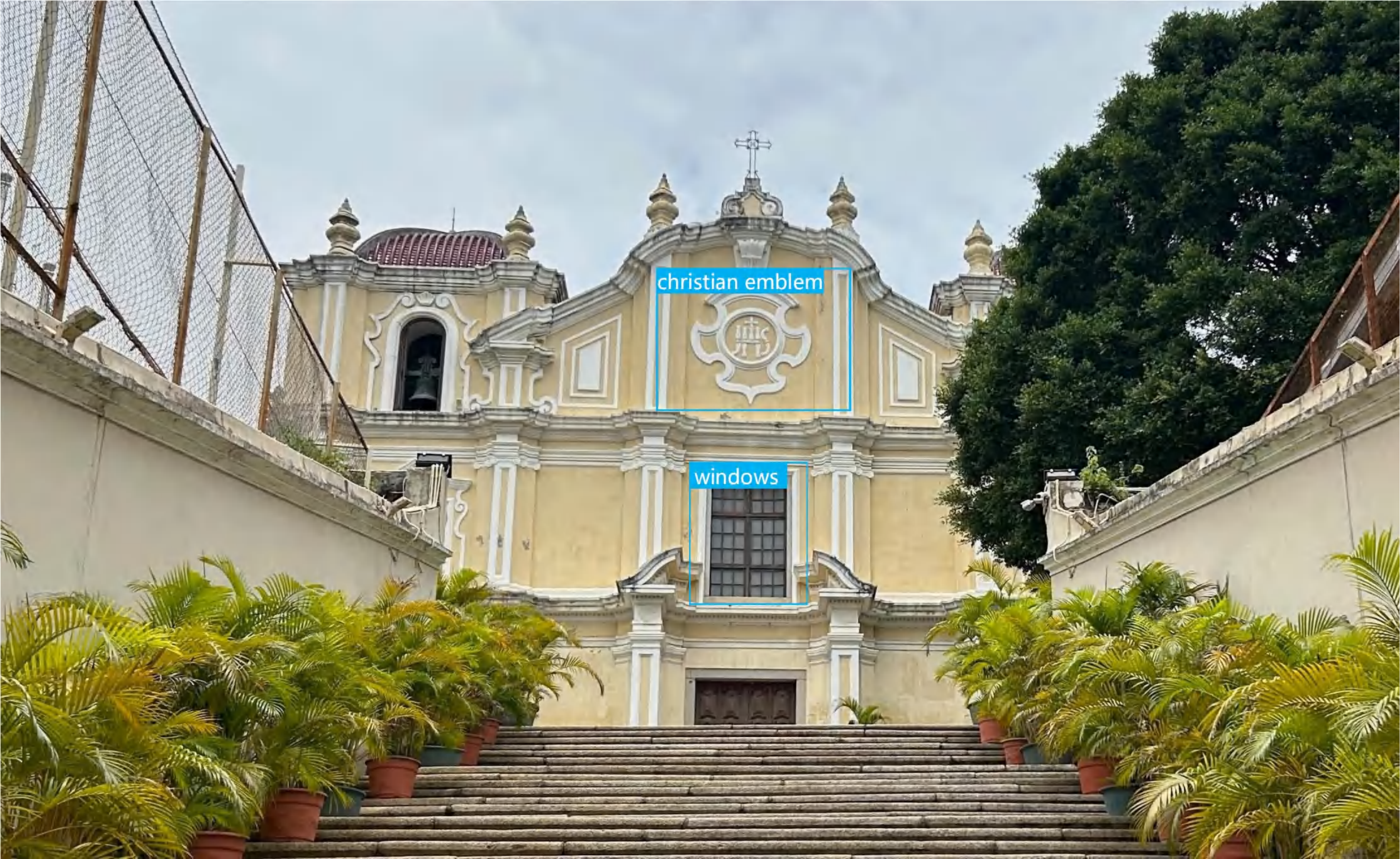} &
\includegraphics[width=\linewidth]{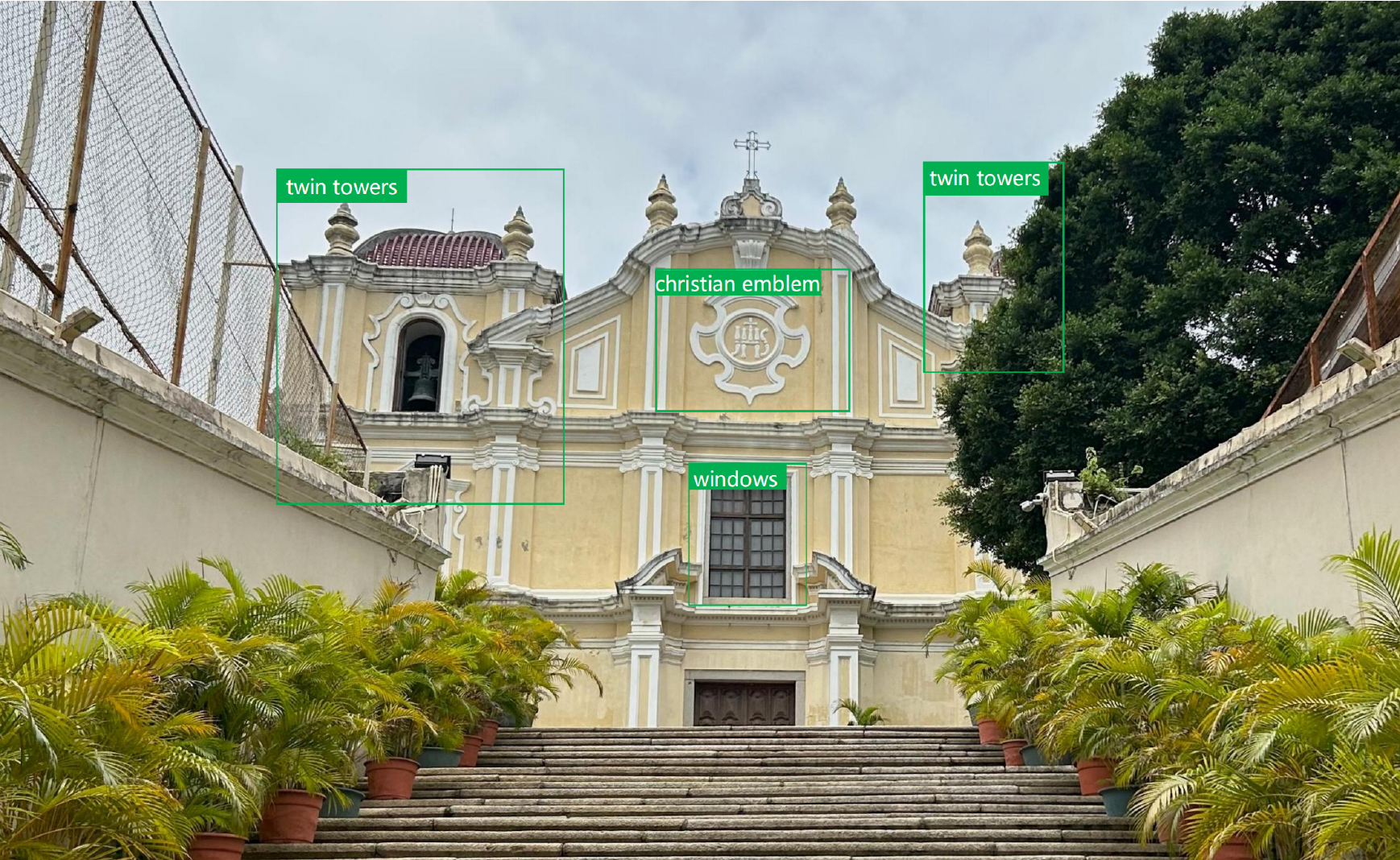} &
\includegraphics[width=\linewidth]{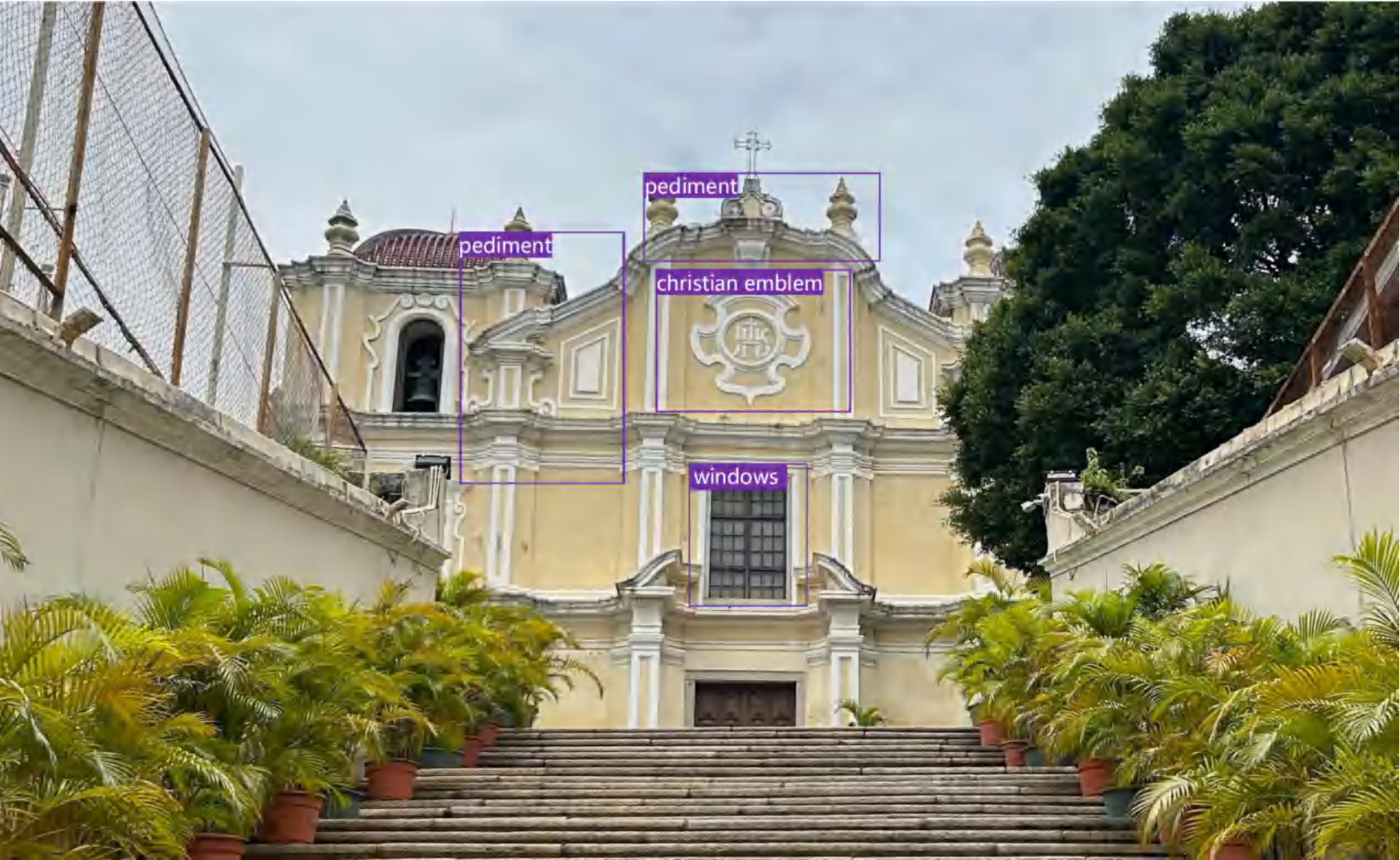} \\

\midrule

\parbox[t]{\hsize}{%
\textbf{Generated Text}\par
The church's main facade presents a massive wall 24.6 meters wide, divided into three vertical and three horizontal sections. \hlBone{The central section features the Christian emblem, with three nails beneath it symbolizing the crucifixion of Jesus.} \hlCone{The second level features three windows: the outer ones are smaller, adorned with arched lintels and surrounding relief carvings, while the central window is a rectangular shuttered opening.} The overall design is simple yet solemn, showcasing the unique charm of religious art.%
} &
\parbox[t]{\hsize}{%
\textbf{Generated Text}\par
The exterior facade of this church displays a majestic Baroque style, spanning a width of 24.6 meters. The symmetrical bell towers at both ends feature glazed tile roofs. Three window openings in the left tower house large bronze bells, while the right tower also holds a bronze bell within its window openings. \hlBtwo{At the building's central pediment, the Jesuit emblem is carved, with the ‘IHS’ above symbolizing the name of Jesus.} \hlCtwo{The second level features three windows: the outer ones are smaller, adorned with arched lintels and surrounding relief decorations, while the central window comprises a large shuttered opening.} The overall design harmoniously blends religious artistry with functionality, conveying the building's solemnity and sacredness.%
} &
\parbox[t]{\hsize}{%
\textbf{Generated Text}\par
The church's main facade presents a massive wall 24.6 meters wide, divided into three vertical and three horizontal sections by two multi-tiered cornices and six fluted pilasters with multiple recesses.\hlGthree{The pediment section of the building displays a quintessential Baroque style, adorned centrally with an exquisite Christian emblem.} \hlBthree{This emblem, rendered in symmetrical geometric patterns and relief carvings, serves as a common religious symbol found in Jesuit churches. Directly beneath the emblem, three nails symbolize the historical crucifixion of Jesus, one of Christianity's most significant emblems. Above the emblem, the letters ``IHS'' derive from the first three letters of Jesus' Greek name ``$\mathrm{IHSOUS}$,'' commonly used as an abbreviation for Jesus Christ, signifying ``Jesus is the Savior of Humanity.''} \hlCthree{The second level features a larger rectangular shuttered window at the center flanked by smaller arched windows. The lintels bear curved ornamentation complemented by sculpted details.} This layout harmonizes with Baroque architecture's pursuit of symmetry and opulence while balancing natural light and aesthetic functionality.%
} \\
\bottomrule
\end{tabular}

\caption{The case study example.}
\label{case study}
\end{table*}

\subsection{Element Description Generation}
\label{app:ed_generation}

For each element $e \in E$, we prompt Qwen3-VL-Max to generate a factual paragraph of one to three sentences describing its visual characteristics, cultural significance, and historical context. We enforce three constraints during generation: factuality requires all statements to be source-grounded with no unverifiable claims; scope constrains the content to the specific element rather than general landmark background; and tone maintains a neutral encyclopedic style that avoids promotional language. Generated paragraphs are stored under their corresponding keys in $\mathrm{ED}$.

\subsection{Validation and Quality Control}
\label{app:validation}

Quality control combines automatic checks with human review. Automatic validation verifies key consistency ($\mathrm{keys}(\mathrm{ED}) = E$), ensures all fields are non-empty, and confirms orthographic normalization. Human reviewers then perform three additional passes: factuality review spot-checks generated descriptions against source materials and regenerates entries containing unsupported claims; style filtering removes promotional language and unattributed superlatives; and terminological harmonization ensures consistent wording for shared elements across landmarks.

\subsection{Ground-Truth Annotation Protocol}
\label{app:annotation}

Human annotators refine LLM-proposed labels $\hat{E}(I)$ to produce $E^{\mathrm{gt}}(I)$ following strict visibility rules. Visual identifiability requires that elements qualify only if they are confidently recognizable from image pixels alone, without relying on prior landmark knowledge; for instance, an external tower invisible in an interior courtyard photo cannot be labeled. For partial occlusion handling, partially occluded elements are included only when discriminative visual cues remain present, such as a half-visible ornamental scroll whose characteristic shape remains recognizable. Synonym resolution requires annotators to select canonical keys when multiple inventory terms refer to near-identical visual components. Uncertain cases undergo ambiguity escalation, where they are marked for expert review or excluded to avoid false positives.

The annotation workflow presents each image $I$ alongside landmark metadata $(\mathrm{meta}, E, \mathrm{ED})$ and LLM-proposed candidates $\hat{E}(I)$. Annotators edit the proposals by adding missed but visible elements, removing hallucinated or non-verifiable elements, and correcting keys to canonical form. This protocol ensures ground-truth labels reflect genuine visual evidence, enabling fair evaluation while accounting for real-world photography ambiguities such as viewpoint limitations and occlusions.

\begin{figure}[t]
  \centering
  \includegraphics[width=0.9\columnwidth]{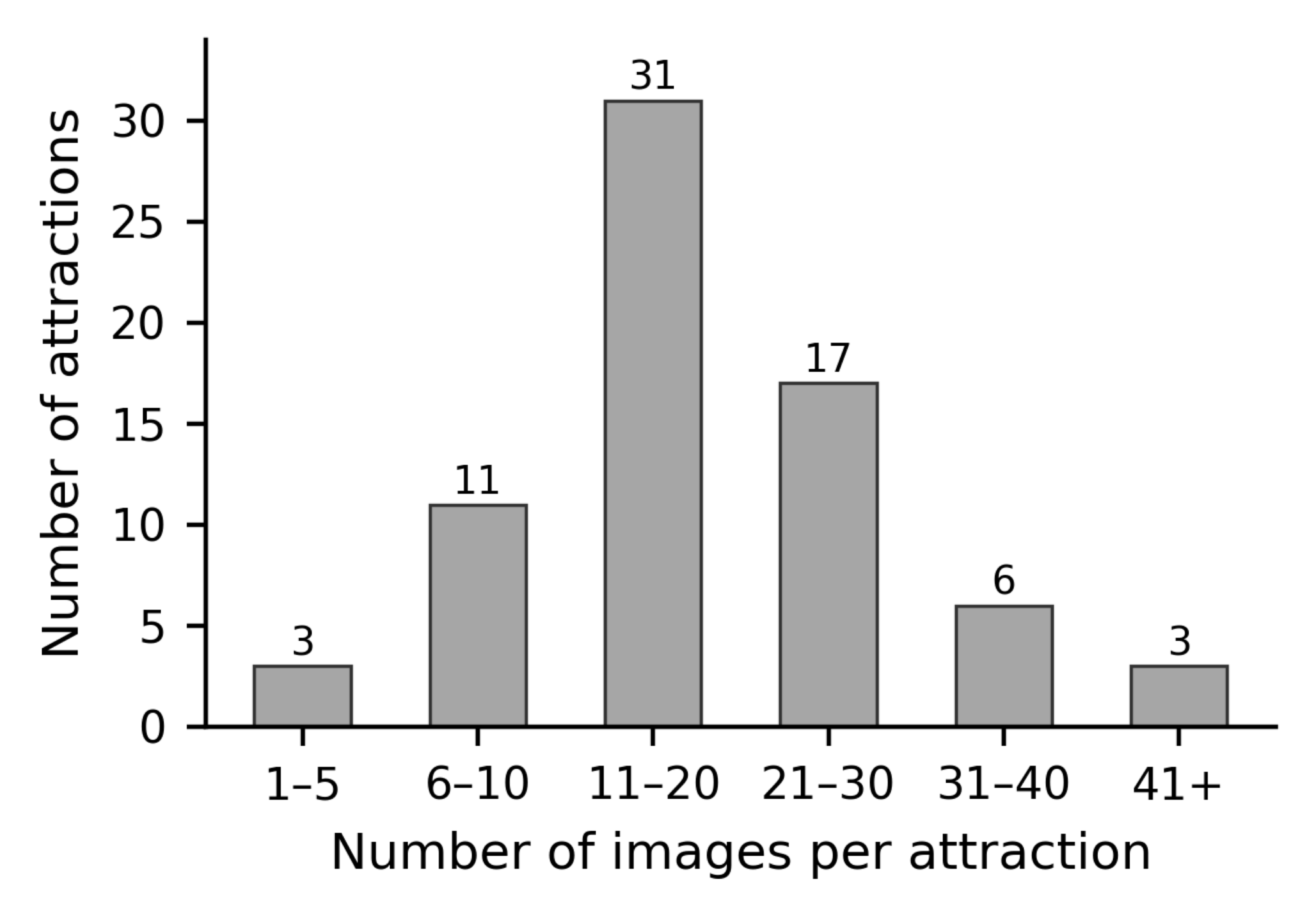}
  \caption{Distribution of image counts per landmark. The dataset shows a single dominant mode around 11--30 images (48 of 71 landmarks, 67.6\%), with only a few landmarks providing more than 40 views.}
  \label{fig:imgcount_dist}
\end{figure}

\subsection{Detailed coverage distribution}
\label{app:statistics}

Beyond the aggregate statistics reported in the main text (mean 34\%, median 29\%), the image distribution statistic is shown in Figure~\ref{fig:imgcount_dist}, which reflects real-world photography patterns: most landmarks contain 11--30 images, while iconic locations exceed 40 images. The element coverage ratio $\text{Coverage}(I) = |E^{\mathrm{gt}}(I)| / |E|$ shows considerable variation. The interquartile range spans 0.18 to 0.47, meaning half of all images show between 18\% and 47\% of available elements. The distribution exhibits notable tail behavior: 12\% of images show fewer than 15\% of elements due to extreme close-up framing, while 5\% show more than 70\% of elements in comprehensive panoramas. Stratifying by view type reveals that close-up views average 22\% coverage, mid-range views average 36\%, and panoramic views average 58\%.

This granular breakdown confirms that multi-view aggregation is essential for our task. No single view type reliably captures complete element inventories, which necessitates cross-view evidence fusion for comprehensive retrieval.

\end{document}